\documentclass[12pt]{article}

\usepackage{color}
\usepackage[utf8]{inputenc}
\usepackage[T1]{fontenc}    % use 8-bit T1 fonts
\usepackage{url}            % simple URL typesetting
\usepackage{booktabs}       % professional-quality tables
\usepackage{amsfonts}       % blackboard math symbols
\usepackage{nicefrac}       % compact symbols for 1/2, etc.
\usepackage{microtype}      % microtypography

\usepackage{amsmath}
\usepackage{amsfonts}
\usepackage{mathrsfs}
\usepackage{amssymb}
\usepackage{amsthm}

\usepackage{thmtools}
\usepackage{thm-restate}

\usepackage{cleveref}

\usepackage{graphicx,subfigure}
\usepackage{tikz}
\usepackage{comment}

\usepackage{epstopdf}

\usepackage{libertine}

\usepackage{algorithm}
\usepackage{algorithmic}

\usepackage[numbers]{natbib}
\newdimen\arrowsize
\pgfarrowsdeclare{arcsq}{arcsq}
{
  \arrowsize=0.2pt
  \advance\arrowsize by .5\pgflinewidth
  \pgfarrowsleftextend{-4\arrowsize-.5\pgflinewidth}
  \pgfarrowsrightextend{.5\pgflinewidth}
}
{
  \arrowsize=1.5pt
  \advance\arrowsize by .5\pgflinewidth
  \pgfsetdash{}{0pt} % do not dash
  \pgfsetroundjoin   % fix join
  \pgfsetroundcap    % fix cap
  \pgfpathmoveto{\pgfpoint{0\arrowsize}{0\arrowsize}}
  \pgfpatharc{-90}{-140}{4\arrowsize}
  \pgfusepathqstroke
  \pgfpathmoveto{\pgfpointorigin}
  \pgfpatharc{90}{140}{4\arrowsize}
  \pgfusepathqstroke
}

\usepackage{chngcntr}
\usepackage{apptools}
\AtAppendix{\counterwithin{theorem}{section}}
\newtheorem{theorem}{Theorem}
\AtAppendix{\counterwithin{lemma}{section}}
\newtheorem{lemma}{Lemma}
\AtAppendix{\counterwithin{equation}{section}}

\title{Transfer Learning with Label Noise}

\font\myfont=cmr12 at 13pt

\author{\myfont Xiyu~Yu\thanks{UBTECH Sydney AI Centre and the School of Information Technologies in the Faculty Engineering and Information Technologies at The University of Sydney, NSW, 2006, Australia, xiyu0300@uni.sydney.edu.au, tongliang.liu@sydney.edu.au, dacheng.tao@sydney.edu.au} \ \ \   Tongliang~Liu\footnotemark[1] \ \ \   
Mingming Gong\thanks{Department of Biomedical Informatics, University of Pittsburgh}~\thanks{Department of Philosophy, Carnegie Mellon University, gongmingnju@gmail.com, kunz1@cmu.edu} \\ \myfont Kun Zhang\footnotemark[3] \ \ \ Kayhan Batmanghelich\footnotemark[2] \ \ \ Dacheng Tao\footnotemark[1]}

\date{}

\begin{document}
\bibliographystyle{plain}

\maketitle

\begin{abstract}
Transfer learning aims to improve learning in target domain by borrowing knowledge from a related but different source domain. To reduce the distribution shift between source and target domains, recent methods have focused on exploring invariant representations that have similar distributions across domains. However, when learning this invariant knowledge, existing methods assume that the labels in source domain are uncontaminated, while in reality, we often have access to source data with noisy labels. In this paper, we first show how label noise adversely affect the learning of invariant representations and the correcting of label shift in various transfer learning scenarios. To reduce the adverse effects, we propose a novel Denoising Conditional Invariant Component (DCIC) framework, which provably ensures (1) extracting invariant representations given examples with noisy labels in source domain and unlabeled examples in target domain; (2) estimating the label distribution in target domain with no bias. Experimental results on both synthetic and real-world data verify the effectiveness of the proposed method.
\end{abstract}

\newpage

\section{Introduction}
In the classical transfer learning setting, given data points $\{x_1^T,\cdots,x_n^T\}$ from the target domain, we aim to learn a function to predict the labels $\{y_1^T,\cdots,y_n^T\}$ using labeled data $\{(x_1^S,y_1^S),
\cdots,(x_m^S,y_m^S)\}$ from a different but related source domain. Let $X$ and $Y$ be the variables of features and labels, respectively. In contrast to the standard supervised learning, the joint distributions $P_{XY}^S$ and $P_{XY}^T$ are different across domains. For example, in medical data analysis, health record data collected from patients of different age groups or hospital locations often vary \cite{purushotham2016variational}. Inferring invariant knowledge from a domain (e.g., an age group or a location) with a large number of observations to another with scare labeled data is desirable since it is laborious to obtain high-quality labels for clinical data \cite{dubois2017effectiveness}. Similarly, in the indoor WiFi localization problem \cite{zhang2013covariate}, the signal distributions received by different phone models are also different. To avoid labeling data for all phone models, it is essential to transfer knowledge from one phone model with sufficient labeled data to another. In this kind of problems, transfer learning techniques can improve the generalization ability of models learned from source domain by correcting domain mismatches.

Due to various assumptions about how the joint distribution $P_{XY}$ shifts across domains, several transfer learning scenarios have been studied. (1) Covariate shift is a traditional scenario where the marginal distribution $P_X$ changes but the conditional distribution $P_{Y|X}$ stays the same. In this situation, several methods have been proposed to correct the shift in $P_X$; for instance, importance reweighting \cite{huang2007correcting} and invariant representation \cite{long2015learning}. (2) Model shift \cite{wang2014active} assumes that the marginal distribution $P_X$ and the conditional distribution $P_{Y|X}$ change independently. In this case, successful transfer requires $Y$ to be continuous, the change in $P_{Y|X}$ to be smooth, and some labeled data to be available in the target domain. (3) Target shift \cite{zhang2013domain} assumes that the marginal distribution $P_Y$ shifts while $P_{X|Y}$ stays the same. In this scenario, $P_X$ and $P_{Y|X}$ will change dependently because their changes are caused by the change in $P_Y$. (4) Generalized target shift \cite{zhang2013domain} assumes that $P_{X|Y}$ and $P_Y$ change independently across domains, causing $P_X$ and $P_{Y|X}$ to change dependently. 
%The difference between these scenarios can be understood from a causal
%perspective \cite{Scholkopf12,zhang2015multi}: (1) and (2) assume that $X$ causes $Y$, while (3) and (4) assume that $Y$ causes $X$. 
An interpretation of the difference between these scenarios from a causal standpoint was also provided \cite{Scholkopf12}. %zhang2015multi

The aforementioned transfer learning methods extract invariant knowledge across different domains based on a strong assumption; that is, the source domain labels are ``clean''. However, it is often violated in practice. This is because that accurately labeling training set tends to be expensive, time-consuming, and sometimes impossible. For example, in medical data analysis, due to the subjectivity of domain experts, insufficient discriminative information, and digitalization errors \cite{saez2016influence}, noisy labels are often inevitable. In computer vision, to reduce the expensive human supervision, we often prefer directly transferring knowledge from easily obtainable but imperfectly labeled source data such as webly-labeled data or machine-labeled data to target data \cite{lee2017cleannet}. 

Therefore, in this paper, we consider the setting of transfer learning that the observed labels in source domain are noisy. The noise is assumed to be random and the flip rates are class-conditional (abbreviated as CCN \cite{natarajan2013learning}), which is a widely-employed label noise model in the machine learning community. The issue is that since we have no access to the true source distribution when the labels are noisy, it might be problematic if we directly apply existing transfer learning methods to correct the mismatches between the noisy source domain and the target domain.

%In this situation, how to extract and transfer
%knowledge from the noisy observations in the source domain becomes even more challenging in transfer learning. To the best of our
%knowledge, this is the first work studying transfer learning with label noise.

As expected, except the covariate shift scenario in which correcting the shift in $P_X$ does not require label information, we can show that label noise can adversely affect most existing transfer learning methods in different scenarios. Taking target shift as an example, in order to correct the shift in $P_Y$, the labeled data
points in the source domain are required to estimate the class ratio between $P^T_Y$ and $P_Y^S$. However,
in the presence of label noise, it is unclear that the class ratio $P^T_Y/P_Y^S$ can be estimated from noisy data. Another example is generalized target shift where $P_Y$ and $P_{X|Y}$ change in an unrelated way. In this scenario, in addition to the possible wrong estimate of $P^T_Y/P_Y^S$, the estimates of invariant representations would be inaccurate because that label noise provides wrong information for matching distributions across domains while learning the representations. Label noise also affects the learning in the model shift scenario, but we will not consider this case because we are concerned with discrete labels and the setting in which there is no label in the target domain.

To address this issue, we propose a label-noise robust transfer learning method in the generalized target shift scenario which is prevalent in transfer learning. To deal with the noisy labels in source domain, we propose a novel method to denoise conditional invariant components. Our method can provably identify the changes in clean distribution $P_Y$, and simultaneously extracts the conditional invariant representations $X'=\tau(X)$ which have similar $P_{X'|Y}$ across domains. Specifically, we construct a new distribution $P^{\textrm{new}}_{X'}$ which is marginalized from the weighted noisy source distribution $P^S_{\rho X',Y}$. Here, we denote $P_\rho$ as the distributions associated with label noise. By matching the distributions $P^{\textrm{new}}_{X'}$ and $P^T_{X'}$, the conditional invariant components and $P^T_Y$  are identifiable from the noisy source data and unlabeled target data. Moreover, in our denoising conditional invariant component framework, we can also theoretically ensure the convergence of the estimate of label distribution in target domain.
%(as shown in Eq. (\ref{eq:marg}))

To verify the effectiveness of the proposed method, we conduct comprehensive experiments on both synthetic and real-world data. The performance are evaluated on classification problems. For fair comparison, after extracting invariant representations using transfer learning methods, we train the robust classifier by employing the forward method in \cite{patrini2017making}. Compared with the state-of-the-art transfer learning methods, our method achieves superior performance. This also indicates that the proposed method is able to transfer invariant knowledge across different domains when label noise is present.

\section{Related Work}
\subsection{Classification with Label Noise}
Learning with noisy labels in classification has been widely studied \cite{long2008random,van2015learning}. These methods can be coarsely categorized into four categories, i.e., dealing with unbiased losses \cite{natarajan2013learning}, label-noise robust losses \cite{khardon2007noise}, label noise cleansing \cite{azadi2015auxiliary}, and label noise fitting \cite{sukhbaatar2014training,patrini2017making}. Similar to many of these methods, we exploit a transition matrix to statistically model the label noise. However, the problem considered in this paper is more challenging because the clean source domain distribution is not assumed to be identical to the target domain distribution. In contrast to classification with label noise, our method can learn transferable knowledge across different domains, where both $P_Y$ and $P_{X|Y}$ may change and the labels of the source data is corrupted. Reports on the general results obtained in this setting are scarce.

\subsection{Traditional Generalized Target Shift Methods}
% From a causal point of view, generalized target shift assumes that the causal mechanism is $Y$ causing $X$. This implies
% that $P_Y$ and $P_{X|Y}$ change independently, while $P_X$ and $P_{Y|X}$ change dependently. Based on this assumption,
% it is possible to identify the changes in $P_Y$ and $P_{X|Y}$ from the changes in $P_X$ under mild conditions.
Existing methods to address generalized target shift usually assume that there exists a transformation $\tau$, e.g., location-scale transformation \cite{zhang2013domain,gong2016domain}, such that the conditional distribution $P_{\tau(X)|Y}$ is invariant across domains. In this paper, we also assume that the conditional invariant components (CICs) exist. We aim to find a transformation $\tau$ such that $P^T(\tau(X)|Y)=P^S(\tau(X)|Y)$ as in \cite{gong2016domain} and to estimate $P^T(Y)$. However, we are given only samples drawn from the distribution $P^T_X$ and the noisy distribution $P^S_{\rho XY}$, which makes the problem more challenging. 

Note that our work is not a simple combination of traditional generalized target shift methods and robust classifiers. As aforementioned, simple combination of transfer learning and label-noise robust classifier overlooks that the knowledge transfer process can be affected by label noise, which thus produces unreliable results. In the setting where only noisy source data and unlabeled target data are available, learning $\tau$ becomes pretty challenging. This is because without clean label $Y$ in both domains, no direct information is available to ensure the matching of conditional densities $P(\tau(X)|Y)$ such that $\tau$ can be learned.  Moreover, it is challenging to estimate $P^T(Y)$ as briefly discussed in the introduction. If $\tau$ is known, the estimation of $P^T(Y)$ is essentially a mixture proportion estimation problem which will be analyzed in the following section. Even if we have the sample from the mixture $P^T(\tau(X))$, the samples of component distributions from source domain are noisy. We cannot obtain correct $P(Y)$ by using methods \cite{gong2016domain,iyer2014maximum}. Therefore, we proposed a novel denoising conditional invariant component framework. It is able to identify $P^T(Y)$ and conditional invariant components $\tau(X)$ from the noisy source data and unlabeled target data. 

In this paper, the simple combinations of transfer learning methods with robust classifiers are included as baselines in our experiments. We show that our method strongly outperforms the baselines, verifying that the superiority of the proposed method to extract invariant knowledge across different domains.

\section{The Effects of Label Noise}

\begin{figure}[t]
{\includegraphics[width=1\columnwidth]{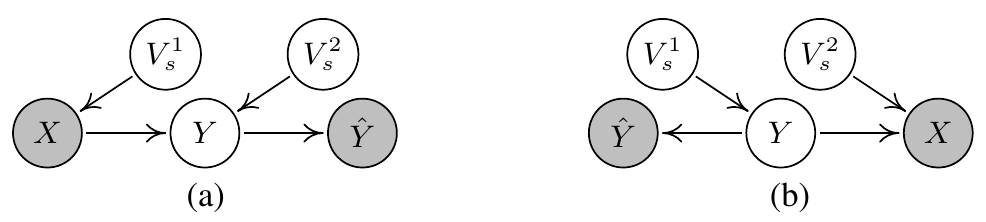}}
\caption{Possible situations of transfer learning with label noise.  $V_s^1$ and $V_s^2$ are independent domain-specific selection variables, leading to changing $P_{XY}$ across domains. (a) Model shift: $V_s^1$ and $V_s^2$ change $P_X$ and $P_{Y|X}$, respectively. (b) Generalized target shift: $V_s^1$ and $V_s^2$ change $P_{Y}$ and $P_{X|Y}$, respectively. In the first scenario, $X$ is a cause for $Y$, whilst in the second scenario, $Y$ is a cause of $X$. If $V_s^2$ is not present, (a) reduces to covariate shift and (b) reduces to target shift. In our setting, the true labels $Y$ in the source domain is unobservable. We only observe noisy labels $\hat{Y}$.}
% (d-f) $Y$ causes $X$. (g) A subset of $X$ causes $Y$, while another subset is an effect of $Y$. }
\label{fig:situation}
\end{figure}

In this section, we examine the effects of label noise in four different transfer learning scenarios, namely 1) covariate shift, 2) model shift, 3) target shift, and 4) generalized target shift. From a causal perspective, 1) and 2) assume that $X$ causes $Y$, indicating that $P_X$ and $P_{Y|X}$ contain no information about each other \cite{scholkopf2012causal}. In transfer learning, the causal relation implies that changes in $P_X$ are independent of changes in $P_{Y|X}$. If the change in $P_{Y|X}$ is large, then it is difficult to correct the shift in $P_{Y|X}$ because we often have no or scarce labels in the target domain. On the contrary, 3) and 4) assume that $Y$ is the cause for $X$, implying that changes in $P_Y$ and $P_{X|Y}$ are independent, while changes in $P_X$ and $P_{X|Y}$ depend on each other. Figure \ref{fig:situation} represents the causal relations between variables in transfer learning using selection diagram defined in \cite{pearl2011transportability}. Here, although the noisy label $\hat{Y}$ is usually generated after $X$ is observed, we exploit the causal model $Y \to \hat{Y}$ according to the assumption that flip rates are independent of features, which is widely employed in the label noise setting \cite{natarajan2013learning,patrini2017making,scott2015rate}.  The effects of label noise in different scenarios are also summarized as follows:

%\begin{itemize}
%\item Covariate shift. 
\textbf{Covariate shift.} In covariate shift \cite{huang2007correcting,zhang2013covariate}, label noise has no effects on the correction of shift in $P_X$. However, after correcting the shift in $P_X$, one needs to take the effects of label noise into account when training a classifier on the source domain \cite{natarajan2013learning,liu2016classification}.

%\item Model shift. 
\textbf{Model shift.} In the model shift scenario \cite{wang2014active}, since $P_X$ and $P_{Y|X}$ change independently, we can correct them separately. Similar to covariate shift, correcting $P_X$ is not affected by label noise. However, correcting shift in $P_{Y|X}$ requires matching $P^S_{Y|X}$ and $P^T_{Y|X}$, which can be seriously harmed by label noise. In this scenario, since a small number of clean labels are assumed to be available in the target domain,  $P_{Y|X}$ is often assumed to change smoothly across domains to reduce the estimation error. The smoothness constraint can reduce the effects of label noise to some extent if one directly matches $P_{\rho Y|X}^S$ and $P_{Y|X}^T$.

\begin{figure}[t!]
\centering
\subfigure []
{\hspace{0pt}\includegraphics[width=0.32\columnwidth]{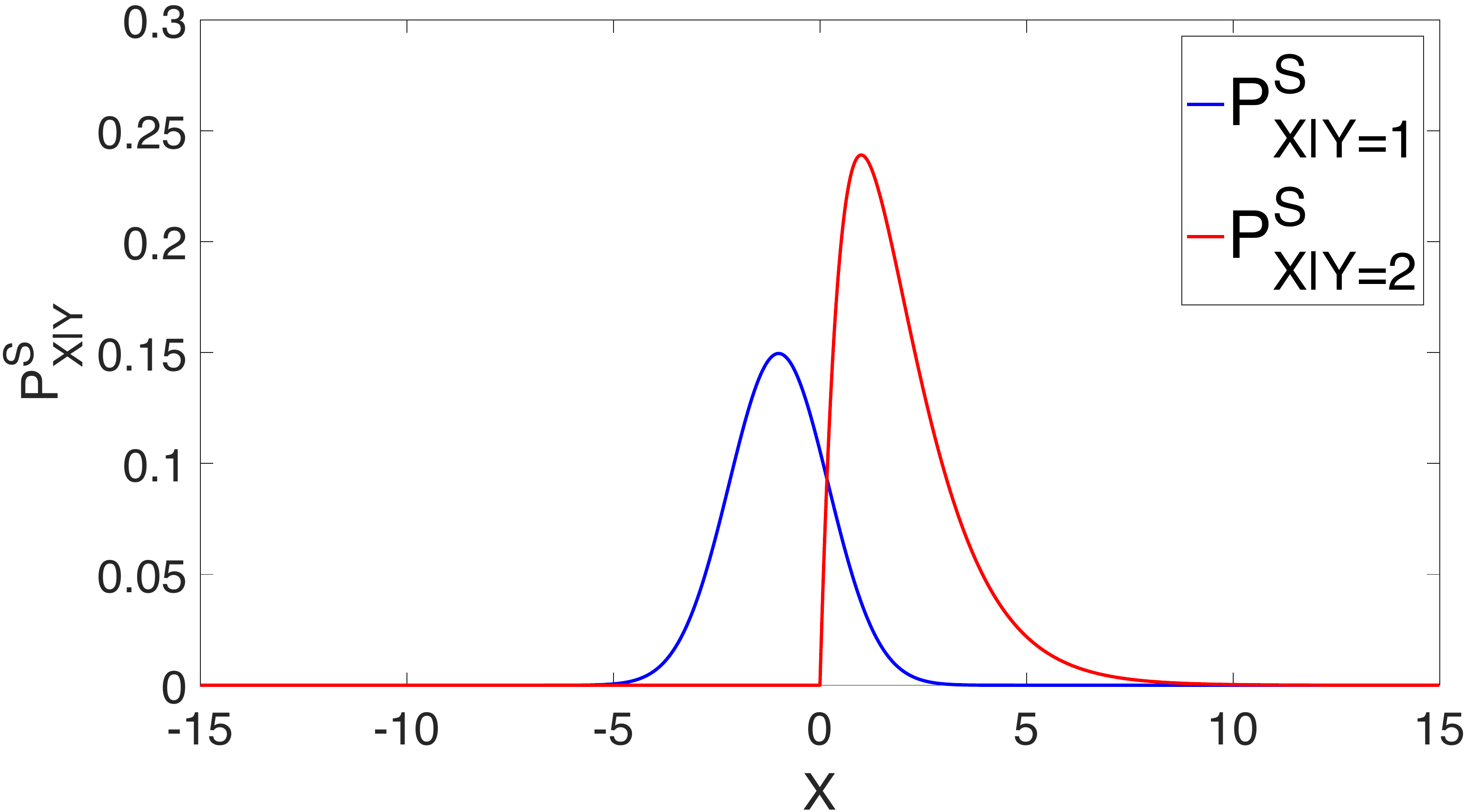}}
\subfigure []
{\hspace{0pt}\includegraphics[width=0.32\columnwidth]{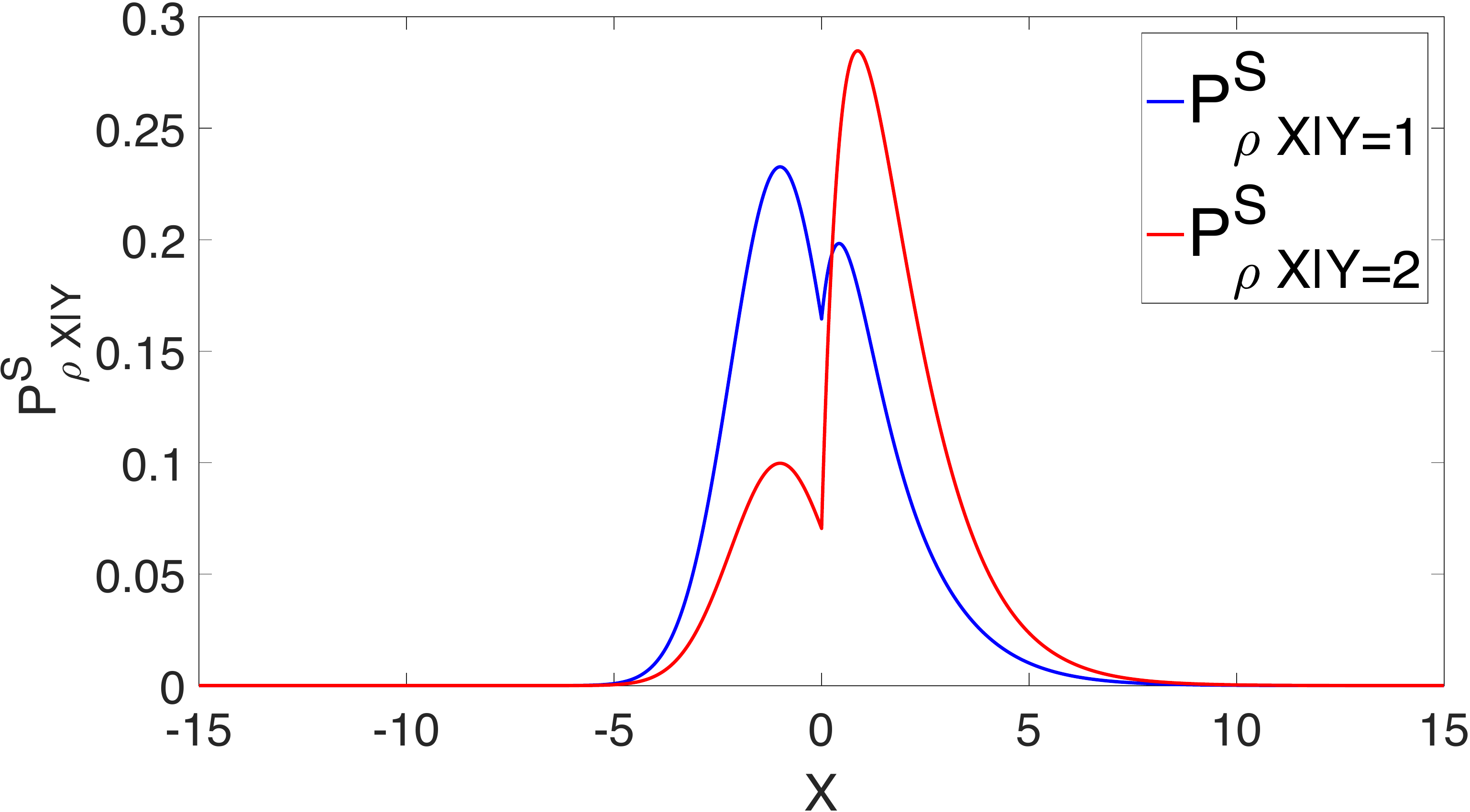}}
\subfigure []
{\hspace{0pt}\includegraphics[width=0.32\columnwidth]{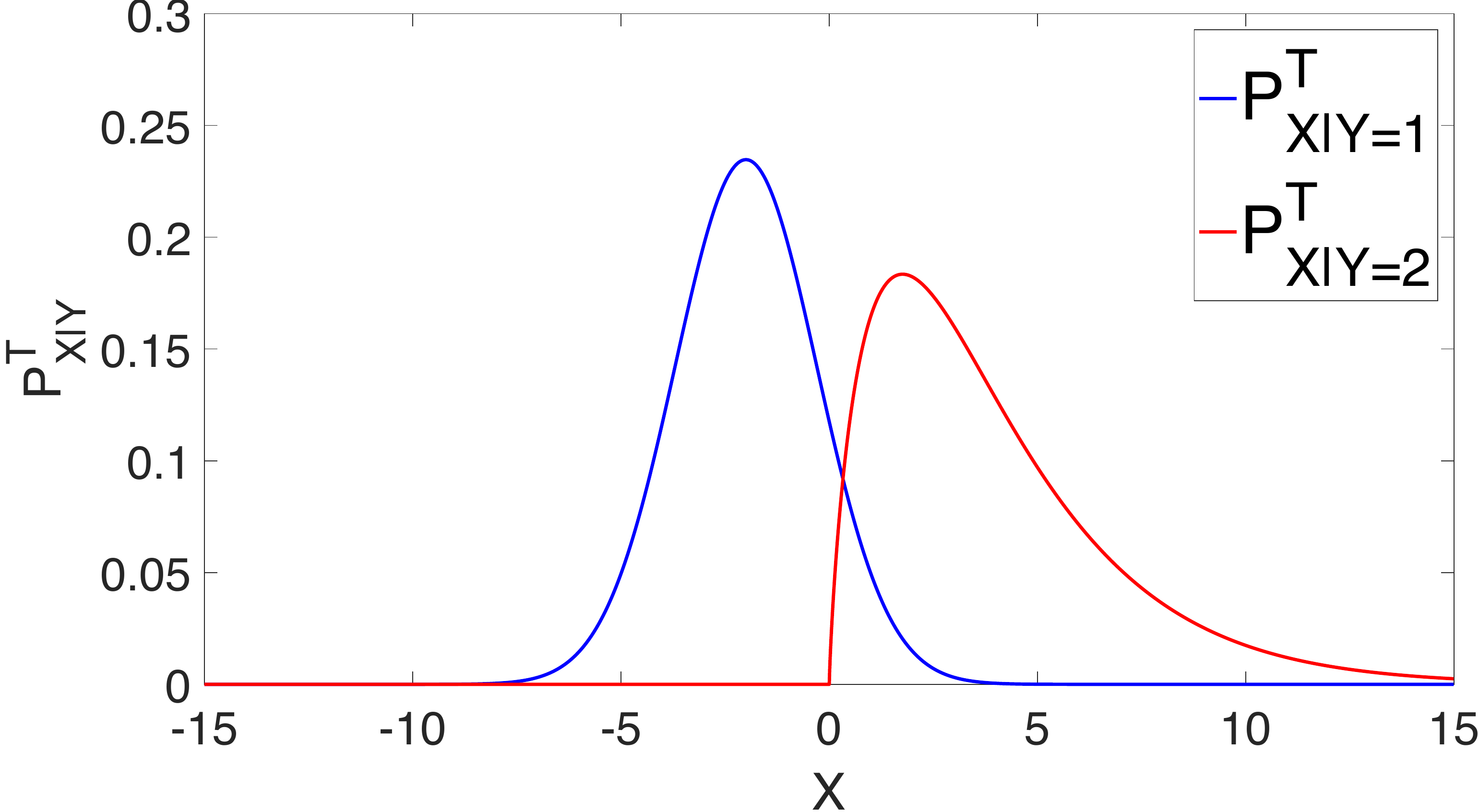}} \label{fig:tarS_noise}
\caption{A simple illustration of the difficulties brought by noisy labels. (a)-(c) show the class conditionals in the clean source domain, noisy source domain, and target domain, respectively. }
\label{fig:ratio_illu}
\end{figure}
%\item Target shift. 
\textbf{Target shift.} In target shift scenario \cite{iyer2014maximum,zhang2013domain}, it is required that $P_{X|Y}^S=P_{X|Y}^T$. The changes in $P_Y$ are often corrected by matching the marginal distribution of the reweighted source domain $P^{\textrm{new}}_X=\sum_{i=1}^c P_{X|Y=i}P^S_{Y=i}\beta(Y=i)$ and the target domain $P^T_X$, where $\beta(Y=i)=P^T_{Y=i}/P^S_{Y=i}$ and $c$ is the class number. 

In the presence of label noise, however, we only have access to $P_{\rho X|Y}^S$ and $P_{\rho Y}^S$ in the source domain. As shown in Figure \ref{fig:ratio_illu},  $P_{\rho X|Y=i}^S$ becomes a mixture of $P_{X|Y=1}^S$ and $P_{X|Y=2}^S$ and is no longer identical to $P^T_{X|Y=i}$. In this case, directly applying the methods in \cite{zhang2013domain,iyer2014maximum} on the noisy data will lead to wrong estimate of $P^T_Y$. Specifically, if we directly employ \cite{zhang2013domain,iyer2014maximum} on noisy data, we need to estimate the mixture proportions for the model $P^T_X=\omega_{\rho 1}P^S_{\rho X|Y=1}+\omega_{\rho 2} P^S_{\rho X|Y=2}$. But the estimated proportions are very likely to be different from those in the mixture model $P^T_X=\omega_1 P_{X|Y=1}+\omega_2 P_{X|Y=2}$. Here, $\omega_i = P^S_{Y=i}\beta(Y=i), i=1,2$.

% In this case, it is unknown whether $P^T_Y$ is still identifiable from the noisy source domain. Directly applying the methods in \cite{zhang2013domain,iyer2014maximum} on the noisy data will lead to wrong results. 
% For example, suppose the clean source domain and target domain share the same distribution, i.e., $P^T_Y=P^S_{Y}$. As the sample size $N\rightarrow \infty$, $\mathbf{1}$, a vector of ones, will be a trivial solution, resulting in $P^T_Y=P^S_{\rho Y}$. However, $P^S_{\rho Y}$ is usually different from $P^S_Y$, leading to a wrong estimate of $P^T_Y$.
Suppose $P^T_Y=P^S_{Y}$ and the label noise is symmetric, i.e., the probability of the labels flipping to each other is the same. Then, it is easy to derive that $(\omega_{\rho 1}, \omega_{\rho 2})$ is the same with $(\omega_{1}, \omega_{2})$. Therefore, as the sample size $m, n\rightarrow \infty$, the estimated density ratio also approaches $\mathbf{1}$, a vector of ones, which is a trivial solution, resulting in $P^T_Y=P^S_{\rho Y}$. However, in most conditions, $P^S_{\rho Y}$ is often different from $P^S_Y$ and label noise is asymmetric, which often leads to a wrong estimate of $P^T_Y$. Thus, we can see the adverse effects of label noise on target shift.

%This is because the presence of label noise violates the causal assumption of independence between changes in $P_Y$ and $P_{X|Y}$ when $Y$ is the cause for $X$. That means, the current transfer learning methods are not able to correct the extra distribution change caused by label noise.

%\item Generalized target shift. 
\textbf{Generalized target shift.} In general target shift \cite{zhang2013domain,gong2016domain}, $P_{X|Y}$ also changes across domains, but it changes independently of $P_Y$. A widely-employed approach is learning conditional invariant components that satisfy $P^S_{X'|Y}=P^T_{X'|Y}$. Under the assumption of conditional invariant components, many works jointly learn $X'$ and $P^T(Y)$ by matching $P^{\textrm{new}}_{X'}=\sum_{i=1}^c P_{X'|Y=i}P^S_{Y=i}\beta(Y=i)$ and $P^T_{X'}$, which naturally requires the information of $P^S_{XY}$ and $P^T_{X}$.

However, in the setting of label noise, similar to target shift, the estimate of invariant components and the label distribution $P^T_Y$ will be inaccurate if we directly use the noisy source distribution $P^S_{\rho XY}$ to correct distribution shift. For example, even though we assume that $X'$ is successfully learned, the estimate of $P^T(Y)$ can be incorrect as that in target shift. A wrong estimate of $P^T(Y)$ can in turn adversely influence the learning of invariant representations in the joint optimization framework \cite{gong2016domain}.

As a conclusion, we can easily observe that label noise is harmful for transferring invariant knowledge and correcting distribution shift in most transfer learning scenarios. We target to reduce these adverse effects of label noise in the following sections.

%{\noindent \em Remainder omitted in this sample. See http://www.jmlr.org/papers/ for full paper.}

% Acknowledgements should go at the end, before appendices and references

%\acks{We would like to acknowledge support for this project
%from the National Science Foundation (NSF grant IIS-9988642)
%and the Multidisciplinary Research Program of the Department
%of Defense (MURI N00014-00-1-0637). }


\section{Label-Noise Robust Transfer Learning}
% New subsection to introduce the uniform framework
% I think the titles of these subsections should be changed.
In this paper, we study a new transfer learning setting in which (1) both distributions $P(X|Y)$ and $P(Y)$ change across different domains; (2) and we are given noisily labeled source data and unlabeled target data. Specifically, denoting $\hat{y}$ as a noisy label, we have access to only ``noisy'' observations $\{(x_1^S,\hat{y}_1^S),\cdots,(x_m^S,\hat{y}_m^S)\}$ in the source domain and unlabeled data $\{x_1^T,\cdots,x_n^T\}$ in the target domain. Here, we consider the class-conditional label noise. The generation of noisy labels is stochastically modeled via a transition probability $P(\hat{Y}=j|Y=i)$, i.e., the flip rate from clean label $i$ to noisy label $j$. All these transition probabilities are summarized into a transition matrix $Q$, where $Q_{ij}=P(\hat{Y}=j|Y=i)$. 

In many transfer learning methods, invariant representation learning and label shift correction is critical for transferring knowledge from source domain to target domain. For example, learning domain-invariant representations are widely-used principles for semantic segmentation \cite{peng2017visda,hoffman2016fcns} and classification \cite{huang2007correcting,gopalan2011domain}. Thus, in this new setting, we also aim to learn the invariant representations and the label distribution $P^T_Y$ in target domain such that the changes in $P(X|Y)$ and $P(Y)$ can be corrected and the effects of label noise can be alleviated.

In the following subsections, we first study how to provably identify invariant representations across different domains and correct the distribution shift in the general target shift scenario with label noise. Then, an importance reweighting framework is introduced for classification problem. Both linear and deep models are finally presented for transfer learning with label noise.

\subsection{Denoising Conditional Invariant Components}

%As indicated in Section 3, in the causal model shown in Figure \ref{fig:situation} (b), $P(X|Y)$ and $P(Y)$ change independently. Thus, we can decouple the correction of these two distributions. As in \cite{gong2016domain}, we first introduce a specific conditional invariant representation which ensures this problem being tractable. That is, we assume that for every $d$-dimensional data $X$, there exists a transformation $\tau: \mathbb{R}^{d} \to \mathbb{R}^{d'}$ satisfying
%\begin{equation}  \label{eq:asummption}
%P^T_{\tau(X)|Y} = P^S_{\tau(X)|Y},
%\end{equation}
%where $X'=\tau(X) \in \mathbb{R}^{d'}$ are the conditional invariant components (CICs) across different domains.

In the label noise setting, learning invariant representations and $P^T_Y$ is very challenging due to the fact that we can only observe the noisy labels but have no information of clean label $Y$ in the source domain. To address this issue, we first introduce a specific conditional invariant representation to ensure this problem being tractable. That is, we assume that for every $d$-dimensional data $X$, there exists a transformation $\tau: \mathbb{R}^{d} \to \mathbb{R}^{d'}$ satisfying
\begin{equation}  \label{eq:asummption}
P^T_{\tau(X)|Y} = P^S_{\tau(X)|Y},
\end{equation}
where $X'=\tau(X) \in \mathbb{R}^{d'}$ are known as conditional invariant ç (CICs) \cite{gong2016domain} across different domains.

Since label noise makes existing transfer learning methods ineffective, we propose a novel method to denoise the conditional invariant components. We find that if the information of label noise model is available, a unique relationship between $P_{\rho}^S(X',Y)$ and $P^T(X')$ can be built, which, in turn, is a clue for us to identify $X'$. 

We observe that label noise does not affect the distribution of $X'$. Then, intuitively, if we marginalize out the variable $\hat{Y}$ of the noisy labels, we may achieve Eq. (\ref{eq:asummption}) by matching the marginal distribution $P_{X'}$. But we need some nontrivial strategies to make it possible. Specifically, we first construct a new distribution $P_{X'}^{\textrm{new}}$, which is marginalized from the reweighted distribution $P^S_{\rho X'Y}$ as follows,
\begin{equation} \label{eq:marg}
P^{\textrm{new}}_{X'} = \sum_{y'} \beta_{\rho}(\hat{Y}=y') P^S_{\rho}(X',\hat{Y}=y') = \sum_y \sum_{y'} \beta_{\rho}(\hat{Y}=y') P^S_{\rho}(X',Y = y, \hat{Y}=y'),
\end{equation}
where $\beta_{\rho}$ are the weights for noisy labels. Note that, in the rest of this paper, when no ambiguity occurs, we use $Y$ as the variable for both ``clean'' and ``noisy'' labels; otherwise, both $Y$ and $\hat{Y}$ are used as variables for ``clean'' and ``noisy'' label, respectively.

Then, under mild conditions, by matching the distribution $P^T_{X'}$ with the new distribution $P_{X'}^{\text{new}}$, we can provably identify the invariant components $\tau(X)$:
\begin{theorem} \label{mainresult}
Suppose the transformation $\tau$ satisfies that $P(\tau(X)|Y=i), i \in \{1,\cdots,c\}$ are linearly independent, and
that the elements in the set $\{v_i P^S(\tau(X)|Y=i)
+ \lambda_i P^T(\tau(X)|Y=i); i \in \{1,\cdots,c\}; \forall v_i, \lambda_i \textrm{ } (v_i^2
+ \lambda_i^2 \neq 0)\}$ are linearly independent.
Then, if $P^{\text{new}}_{X'}=P^T_{X'}$, we have $P^T_{X'|Y} = P^S_{X'|Y}$; and
$\beta(Y=y)=\sum_{y'} P^{S}(\hat{Y}=y'|Y=y)\beta_{\rho}(\hat{Y}=y'),
\forall y, y' \in \{1,\cdots,c\}$, where $\beta(Y=y)=P^T(Y=y)/P^S(Y=y)$.
\end{theorem}

Please see the proof of Theorem \ref{mainresult} in Appendix A. Note that the linearly independent property is a weak assumption which has been widely used as the basic condition for class ratio estimation \cite{gong2016domain} and mixture proportion estimation \cite{yu2018efficient}. 

Let $\mathbf u = [\beta(Y=1),\cdots,\beta(Y=c)]^{\top}$ and ${\mathbf u}_{\rho} = [\beta_{\rho}(Y=1),\cdots,\beta_{\rho}(Y=c)]^{\top}$. According to Theorem \ref{mainresult}, we have $\mathbf u = Q {\mathbf u}_{\rho}$. In label noise, we often assume that $Q$ is usually diagonally dominant and invertible. Then, the relationship between $\beta_{\rho}$ and $\beta$ is uniquely determined, as well as the relationship between $P_{\rho}^S(X',Y)$ and $P^T(X')$. In this case, if $Q$ is known and these two marginal distributions are successfully matched, we can (1) identify the conditional invariant components; (2) and learn $\beta_{\rho}$ which indicates that the changes in the distribution $P_{Y}$ is also identifiable. In practice, the transition matrix $Q$ is not available, but we can usually estimate it by methods in \cite{liu2016classification,patrini2017making}.

\subsubsection{Denoising MMD Loss} 
To enforce the matching between $P^{\text{new}}_{X'}$ and $P^{T}_{X'}$, we employ the kernel mean matching of these two distributions and minimize the squared maximum mean discrepancy (MMD) loss:
\begin{equation} \label{main}
\|\mu_{P^{\text{new}}_{X'}}[\psi(X')]-\mu_{P^{T}_{X'}}[\psi(X')]\|^2 = \|\mathbb{E}_{X'\sim P^{\text{new}}_{X'}}[\psi(X')]-\mathbb{E}_{X'\sim P^{T}_{X'}}[\psi(X')]\|^2,
\end{equation}
where $\psi$ is a kernel mapping.
% It is easy to verify that $\mathbb{E}_{(X',Y) \sim P^S_{\rho X'Y}}[\beta_{\rho}(Y)\psi(X')] =
% \mathbb{E}_{(X',Y) \sim P^S_{X'Y}}[\beta(Y)\psi(X')]$.
According to Eq. (\ref{eq:marg}), we have
\[\mathbb{E}_{X'\sim P^{\text{new}}_{X'}}[\psi(X')]
= \mathbb{E}_{(X',Y) \sim P^S_{\rho X'Y}}[\beta_{\rho}(Y)\psi(X')].
\]
Therefore, minimizing Eq. (\ref{main}) is equivalent to minimizing
\[\|\mathbb{E}_{(X',Y) \sim P^S_{\rho X'Y}}[\beta_{\rho}(Y)\psi(X')] -
\mathbb{E}_{X'\sim P^{T}_{X'}}[\psi(X')]\|^2.\]

In practice, we can only observe the corruptly labeled source data $\{(x_1,\hat{y}_1^S),\cdots,(x_m,\hat{y}_m^S)\}$ and the unlabeled target data $\{x_1^T,\cdots,x_n^T\}$. Therefore, we approximate the expected kernel mean values by the empirical ones:
\begin{equation} \label{eq:obj}
% \begin{aligned}
\|\frac{1}{m}\psi(\mathbf{x'}^{S})\beta_{\rho}(\mathbf{\hat{y}}^{S})-
\frac{1}{n}\psi(\mathbf{x'}^{T})\mathbf{1}\|^2,\\
\end{equation}
where $\beta_{\rho}(\mathbf{\hat{y}}^{S}) = [\beta_{\rho}(\hat{y}_1),\cdots,\beta_{\rho}(\hat{y}_{m})]^{\top}$;
$\mathbf{x'}$ denotes the matrix of the invariant representations.

%\textbf{Reparametrization of $\beta_{\rho}(\mathbf{\hat{y}}^{S})$.} 
However, Eq. (\ref{eq:obj}) is not explicitly formulated w.r.t. $P^T_Y$. If we directly optimizing Eq. (\ref{eq:obj}) w.r.t. $\beta_{\rho}(\mathbf{\hat{y}}^{S})$, it will result in incorrect $\beta_{\rho}$ that violates the fact that $\beta_{\rho}(\hat{y})$ should be the same for the same $\hat{y}$. It is thus impossible to identify $P^T_Y$.

Therefore, we need to reparameterize the formulation by applying the relationship between $\beta_{\rho}$ and $P^T_Y$ in Theorem \ref{mainresult}, i.e., $\beta_{\rho}(\hat{Y}=i) = \sum_{j=1}^c Q^{-1}_{ij}\frac{P^T(Y=j)}{P^S(Y=j)}$. It is also easy to derive that 
$[P^S(Y=1),\cdots,P^S(Y=c)] Q = [P_\rho^S(Y=1),\cdots,P_\rho^S(Y=c)]$. Given estimated $\hat{Q}$
and $[\hat{P}_\rho^S(Y=1),\cdots,\hat{P}_\rho^S(Y=c)]^\top$, we can construct the vectors $\mathbf g_i = [\frac{\hat{Q}^{-1}_{i1}}{\hat{P}^S(Y=1)},
\cdots,\frac{\hat{Q}^{-1}_{ic}}{\hat{P}^S(Y=c)}], i \in \{1,\cdots,c\}$. If $\hat{y}_k=i$, $\forall k \in \{1,\cdots,m\}$, define the matrix $G \in \mathbb{R}^{m\times c}$,
where the $k$-th row of $G$ is $\mathbf g_i$.
Let $\beta_{\rho}(\mathbf{\hat{y}}^{S})=G\mathbf{\alpha}$. Then, $\alpha$ is an estimate of 
$[P^T(Y=1),\cdots,P^T(Y=c)]^{\top}$.

The denoising MMD loss now can be reparametrized as
\begin{equation} \label{linear_model}
\|\frac{1}{m}\psi(\mathbf{x'}^{S})G\alpha-\frac{1}{n}\psi(\mathbf{x'}^{T})\mathbf{1}\|^2 = \frac{ \alpha^{\top}G^{\top} \mathbf{K}^S G \alpha}{m^2}- \frac{2\mathbf{1}^{\top} \mathbf{K}^{T,S} G \alpha}{mn} + \frac{\mathbf{1}^{\top} \mathbf{K}^{T} \mathbf{1}}{n^2},
\end{equation}
where $\mathbf{K}^S$ and $\mathbf{K}^T$ are the kernel matrix of $\mathbf x'^S$ and $\mathbf x'^T$, respectively; $\mathbf{K}^{T,S}$ is the cross kernel matrix. In this paper, the Gaussian kernel, i.e., $k(x_i,x_j)=\exp{\left(-\frac{\|x_i-x_j\|^2}{2\sigma^2}\right)}$ is applied, where $\sigma$ is the bandwidth. 

Therefore, according to Theorem \ref{mainresult}, optimizing the denoising MMD loss in Eq. (\ref{linear_model}) ensures us to identify the conditional invariant components and $P^T(Y)$.

\subsection{Importance Reweighting} 
Since the denoising MMD loss can provably identify conditional invariant components and correct label shift, we can now learn label-noise robust classifiers. In this classification problem, we aim to learn a hypothesis function $f^*: \mathbb{R}^{d'}\to \mathbb{R}^c$ from the noisy source data that can generalize well on the target data. Ideally, $f^*$ minimizes the expected loss $\mathbb{E}_{(X',Y)\sim P^T_{X'Y}} [\ell(f(X'),Y)]$, where $\ell$ is the loss function; $X'=\tau(X)$ are the conditional invariant components of $X$.

In practice, we often assume that $f^*$ can predicts $P^T(Y|X')$ \cite{reid2010composite,patrini2017making} and $\arg\max_{i\in\{1,\cdots,c\}} f^*_i$ predicts the label. Here, $f^*_i$ is the $i$-th entry of $f^*$. To facilitate the learning of $f^*$, we first imagine that the target domain has the same label noise model as the source domain. Note that, this does not necessarily imply that label noise really exists in target domain because, in our setting, we even have no label information of target data. We can see, the minimizer $f_{\rho}^*=\arg\min_f\int\ell(f(X'),Y)P_{\rho}^T(X',Y)dX'dY$ is also assumed to be able to predict $P_{\rho}^T(Y|X')$. If the classifier $f_{\rho}^*$ is found and $Q$ is invertible, we can obtain $f^*$ according to the following relationship:
 \begin{equation} \label{classifer_relation}
 [P^T(Y=1|X'),\cdots,P^T(Y=c|X')] Q =[P_\rho^T(Y=1|X'),\cdots,P_\rho^T(Y=c|X')].
 \end{equation}

Thus, the problem remains to learn $f_{\rho}^*$, which can be obtained by exploiting the importance reweighting strategy: 
\begin{align*}
&f_{\rho}^*=\arg\min_f\int\ell(f(X'),Y)P_{\rho}^T(X',Y)dX'dY\\
&=\arg\min_f\int\frac{P_{\rho}^T(X',Y)}{P_{\rho}^S(X',Y)}\ell(f(X'),Y)P_{\rho}^S(X',Y)dX'dY.
\end{align*}

Since $P^T_{\rho}(X',Y)$ is constructed from $P^T(X,Y)$ by using the same transition matrix $Q$ and $P^T(X'|Y)=P^S(X'|Y)$, we can easily have $P_\rho^T(X'|Y)=P_\rho^S(X'|Y)$ and thus
\begin{align*}
f_{\rho}^*&=\arg\min_f\int\frac{P_{\rho}^T(Y)}{P_{\rho}^S(Y)}\ell(f(X'),Y)P_{\rho}^S(X',Y)dX'dY \\
&=\arg\min_f\int \gamma(Y) \ell(f(X'),Y)P_{\rho}^S(X',Y)dX'dY,
\end{align*}
where $\gamma(Y)=\frac{P_{\rho}^T(Y)}{P_{\rho}^S(Y)}$. In practice, only the training sample is observable, we thus minimize the empirical loss,
\begin{equation} \label{main_framework}
\hat{R} = \frac{1}{m} \sum_{i=1}^m \gamma(\hat{y}^S_i) \ell(f(x'^S_i), \hat{y}_i^S),
\end{equation} 
to find the approximated classifier $f_{\rho}$. 

Instead of separately finding $f^*_{\rho}$ by minimizing Eq. (\ref{main_framework}) and transiting $f^*_{\rho}$ to $f^*$ according to Eq. (\ref{classifer_relation}), in this paper, we employ the forward strategy proposed in \cite{patrini2017making}; that is, we directly minimize the following risk,
\begin{equation} \label{final_objective}
\hat{R} = \frac{1}{m} \sum_{i=1}^m \gamma(\hat{y}^S_i) \ell(Q^{\top} f(x'^S_i), \hat{y}_i^S),
\end{equation}
As we know, by minimizing the empirical risk in Eq. (\ref{final_objective}), $Q^{\top} f(x'^S_i)$ can approximately predict $P^T_{\rho}(Y|X')$. Then, according to Eq. (\ref{classifer_relation}), $f(x'^S_i)$ can finally approximately predict $P^T(Y|X')$.

Note that, in practice, the ratio $\gamma(Y)$ is also unknown. But $P_{\rho}^S(Y)$ can be empirically estimated from the noisy source data, and $P^T(Y)$ is estimated by our denoising MMD loss, $P_{\rho}^T(Y)$ can also be computed according to the relationship similar to Eq. (\ref{classifer_relation}). In this way, $\gamma(Y)$ can be obtained.

\subsection{The Overall Models}
We are now ready to introduce the proposed models. In order to extract conditional invariant components, the transformation $\tau$ varies from linear ones to non-linear ones depending on the complexity of input data space. We accordingly propose the following two representative transfer learning models.

\subsubsection{Linear Model} 
Linear model is a two-stage model in which we first identify invariant representations and $P^T(Y)$ and then train the classifier according to the importance reweighting framework. In linear model, $\tau(x_i)=x'_i=W^{\top}x_i$. To avoid the trivial solution, $W$ is constrained to be orthogonal. Then, according to Eq. (\ref{linear_model}), we have
\begin{equation}
\begin{aligned}
&\min_{W,\alpha} \hat{\mathcal{D}}(W,\alpha)=\|\frac{1}{m}\psi(W^{\top}\mathbf{x}^{S})G\alpha-
\frac{1}{n}\psi(W^{\top}\mathbf{x}^{T})\mathbf{1}\|^2,\\  
&\text{s.t.}  \quad W^{\top}W = I; \sum_{i=1}^c \alpha_i = 1; \\
&\quad \quad \alpha_i\geq 0, \forall i \in \{1,\cdots,c\}.\nonumber
\end{aligned}
\end{equation}

Note that even though the objective function has similar form with that in \cite{gong2016domain}, it is essentially different. This is because in this objective function, the source data is noisily labeled and $G$ is carefully designed to relate $P^S_{\rho}(X,Y)$ and $P^T(X)$ such that conditional invariant components and $P^T(Y)$ can be identified from noisy source data and unlabeled target data.

The alternating optimization method is applied to update $W$ and $\alpha$. Specifically, we apply the conjugate gradient algorithm on the Grassmann manifold to optimize $W$, and use the quadratic programming to optimize $\alpha$. After identifying the invariant representations and $P^T_Y$ by solving above problem, we can then use them to train a classifier for the target data by minimizing Eq. (\ref{final_objective}).

\begin{figure}[t]
\begin{center}
{\includegraphics[width=0.8\columnwidth]{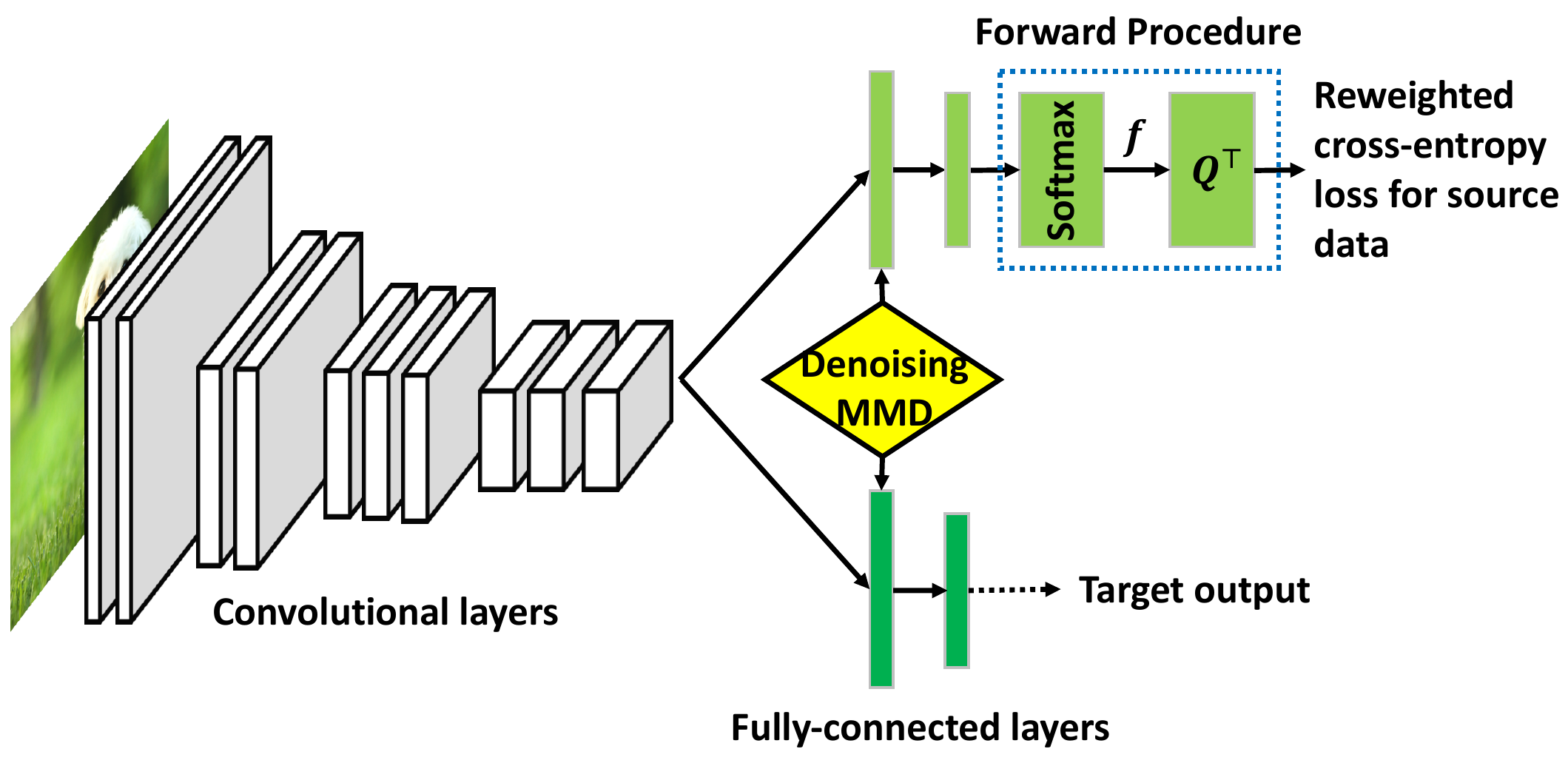}}
\caption{An overview of the proposed end-to-end deep transfer learning model.} \label{fig_overview}
\end{center}
\end{figure}

\subsubsection{Deep Model} 
Besides the two-stage linear model, we also propose an end-to-end learning model incorporating deep neural networks, which have been proven to be effective to extract invariant knowledge across different domains \cite{long2015learning,long2016deep}. Here, we modify the conventional deep neural network for classification, e.g., AlexNet \cite{krizhevsky2012imagenet}, in two aspects: (1) Due to the fact that the domain discrepancy becomes larger for the features in higher layers, we impose the denoising MMD loss on a higher layer for extracting the invariant representations; (2) to learn a classifier robust to label noise, we also add the forward procedure \cite{patrini2017making} before the cross-entropy (CE) loss as in Eq. (\ref{final_objective}). The structure is shown in Figure \ref{fig_overview}.

Specifically, let $h^{l}$ be the responses of the $l$-th hidden layer, $W_{1:l}$ be the parameters in the $1$-th to $l$-th layers, and $L$ be the total number of layers in the deep neural network. Suppose that we impose the denoising MMD loss on the features in the $l_1$-th layer; that is, $\tau(x_i)=h_i^{l_1}$. Then, the denoising MMD loss is
\begin{equation}
\hat{D}(W_{1:l_1},\alpha) = \|\frac{1}{m}\psi({\mathbf{h}^{l_1}}^{S})G\alpha-
\frac{1}{n}\psi({\mathbf{h}^{l_1}}^{T})\mathbf{1}\|^2,
\end{equation} 
where ${\mathbf{h}^{l_1}}$ is the matrix of the responses of the $l_1$-th layer. 

Denote $f(x_k)$ as the softmax output w.r.t. the input $x_k$ (see Figure \ref{fig_overview}). According to Eq. (\ref{final_objective}), the loss for classification is
%\begin{equation}
%\hat{J}(W_{1:L}) = \frac{1}{m} \sum_{k=1}^m \gamma(\hat{y}_k) CE(Q^{\top} g(x_k), \hat{y}_k),
%\end{equation}
%where $\gamma(\hat{y}_k)=\frac{\alpha^{\top}Q_{:i}}{P^S(Y=i)}$ if $\hat{y}_k = i$. $Q_{:i}$ is the $i$-th column of $Q$. The detailed derivation of $\gamma(\hat{y}_k)$ can be found in the Supplementary Material. Together with the regularization on the parameters $\Omega(W_{1:L})$ (e.g., $l_2$ norm), we can obtain our proposed model:
\begin{equation}
\hat{R}(W_{1:L}) = \frac{1}{m}  \sum_{k=1}^m \gamma(\hat{y}^S_k) CE(Q^{\top} f(x^S_k), \hat{y}^S_k),
\end{equation}
where $\gamma(\hat{y}^S_k)=\frac{\alpha^{\top} Q_{:i}}{P^S_{\rho}(Y=i)}$ if $\hat{y}^S_k = i$; $Q_{:i}$ denotes the $i$-th column of $Q$. Together with the regularization $\Omega(W_{1:L})$ (e.g., $l_2$ norm) of the parameters, our final model becomes
\begin{equation} \label{main_obj}
\begin{aligned}
\min_{W_{1:L},\alpha} &\hat{R}(W_{1:L})+\pi_1 \hat{D}(W_{1:l_1},\alpha)+\pi_2 \Omega(W_{1:L}), \\
\textrm{s.t. } & \sum_{i=1}^c \alpha_i = 1; \alpha_i \geq 0, \forall i \in \{1,\cdots,c\},
\end{aligned}
\end{equation}
where $\pi_1$ and $\pi_2$ are the tradeoff parameters of denoising MMD loss and regularization, respectively. Again, by minimizing Eq. (\ref{main_obj}), if $Q^{\top} f(X)$ approximates $P^T_{\rho}(Y|X)$, then $f(X)$ approximates $P^T(Y|X)$. We can then successfully learn the classifier for the target data.

\subsection{Convergence Analysis}\label{analysis}
In this subsection, we study the convergence rates of the estimators to the true label noise rates and optimal class priors. The convergence rate for estimating the label noise rates has been well studied under the ``anchor set'' condition that for any $y$ there exist $x$ in the domain of $X$ such that $P(Y=y|X)=1$ and $P(Y=y'|X)=0, \forall y'\neq y$, which is likely to be held in practice. For example, two estimators with convergence guarantees has been proposed in \cite{liu2016classification} and \cite{scott2015rate}, respectively. Recently, \cite{ramaswamy2016mixture} exploited the ``anchor set'' condition in Hilbert space and designed estimators that can converge to the true label noise rates with an order of $O(m^{-\frac{1}{2}})$. Some work based on a weaker assumption, i.e, linearly independent assumption, is also proposed to estimate label noise, and a fast convergence is also guaranteed \cite{yu2018efficient}. Therefore, we mainly focus on the convergence analysis of estimating class ratios.

In order to analyze the convergence rate of the estimated class prior $\hat{\alpha}$ to the optimal $\alpha^*$ in the presence of label noise, we first abuse the training samples $\{(x_1^S,\hat{y}_1^S),\cdots,(x_m^S,\hat{y}_m^S)\}$ and $\{x_1^T,\cdots,x_n^T\}$  as i.i.d. variables, respectively. Abuse $W$ as the parameters related to the transformation $\tau$ and
\[\mathcal{D}({W},{\alpha})=\|\mathbb{E}\frac{1}{m}\psi(\mathbf{x}'^{S})G\alpha-\mathbb{E}\frac{1}{n}\psi(\mathbf{x}'^{T})\mathbf{1}\|^2.\]
We analyze the convergence rate by deriving an upper bound for $\mathcal{D}({W},\hat{\alpha})-\mathcal{D}({W},{\alpha}^*)$ with fixed $Q$ and $W$.
\begin{theorem} \label{maintwo}
Given learned $\hat{Q}$ and $\hat{W}$, let the induced RKHS be universal and upper bounded that $\|\psi(\tau(x))\|\leq\wedge_{\hat{W}}$ for all $x$ in the source and target domains, and let the entries of $G$ be bounded that $|G_{ij}|\leq\wedge_{\hat{Q}}$ for all $i\in \{1,\cdots,m\}, j\in\{1,\cdots,c\}$. $\forall \delta>0$, with probability at least $1-\delta$, we have
\begin{equation}
\mathcal{D}(\hat{W},\hat{\alpha})-\mathcal{D}(\hat{W},\alpha^*)\leq8(\wedge_{\hat{Q}}+1)^2\wedge_{\hat{W}}^2\sqrt{\frac{\sqrt{c}}{\sqrt{m}}+\frac{\sqrt{c}}{\sqrt{n}}+\sqrt{2(\frac{1}{m}+\frac{1}{n})\log\frac{1}{\delta}}}.
\end{equation}
\end{theorem}
See the proof of Theorem \ref{maintwo} in Appendix B. Although the bound in Theorem \ref{maintwo} involves two fixed parameters, the result is informative if $Q^*$ and $W^*$ are given or $\hat{Q}$ and $\hat{W}$ quickly converges to $Q^*$ and $W^*$, respectively. From previous analyses, we know that fast convergence rates for estimating label noise rate are guaranteed. However, the convergence of $\hat{W}$ to $W^*$ is not guaranteed because the objective function is non-convex w.r.t. $W$. How to identify the transferable components $\tau(X)$ should be further studied.

\section{Experiments}
To show the robustness of our method to label noise, we conduct comprehensive evaluations on both simulated and real data. We first compare our method, denoising conditional invariant components (abbr. as DCIC hereafter), with CIC \cite{gong2016domain} on identifying the changes in $P_Y$ given noisy observations. The effectiveness of the linear and deep models is then verified on both the synthetic and real data. We compare DCIC with the domain invariant projection (DIP) method \cite{baktashmotlagh2013unsupervised}, transfer component analysis (TCA) \cite{pan2011domain}, Deep Adaptation Networks (DAN) \cite{long2015learning} and CIC \cite{gong2016domain}. In all experiments, the bandwidth $\sigma$ of the Gaussian kernel is set to be the median value of the pairwise distances between all raw features (linear model) or between all the extracted invariant features (deep model).

\subsection{Synthetic Data}
\begin{figure}[t!]
\centering
\subfigure []
{\hspace{-0pt}\includegraphics[width=0.32\columnwidth]{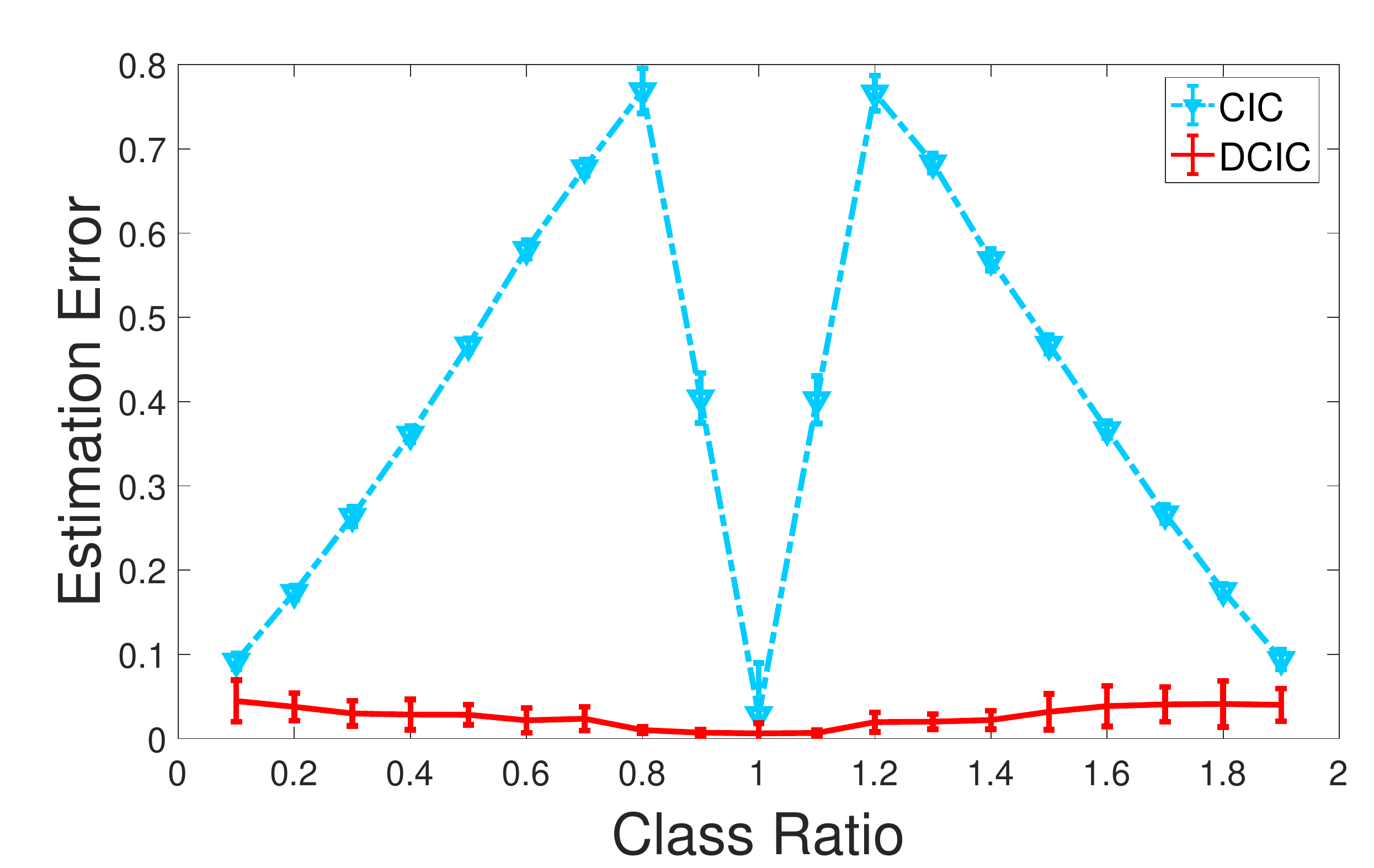}}
\subfigure []
{\hspace{-0pt}\includegraphics[width=0.32\columnwidth]{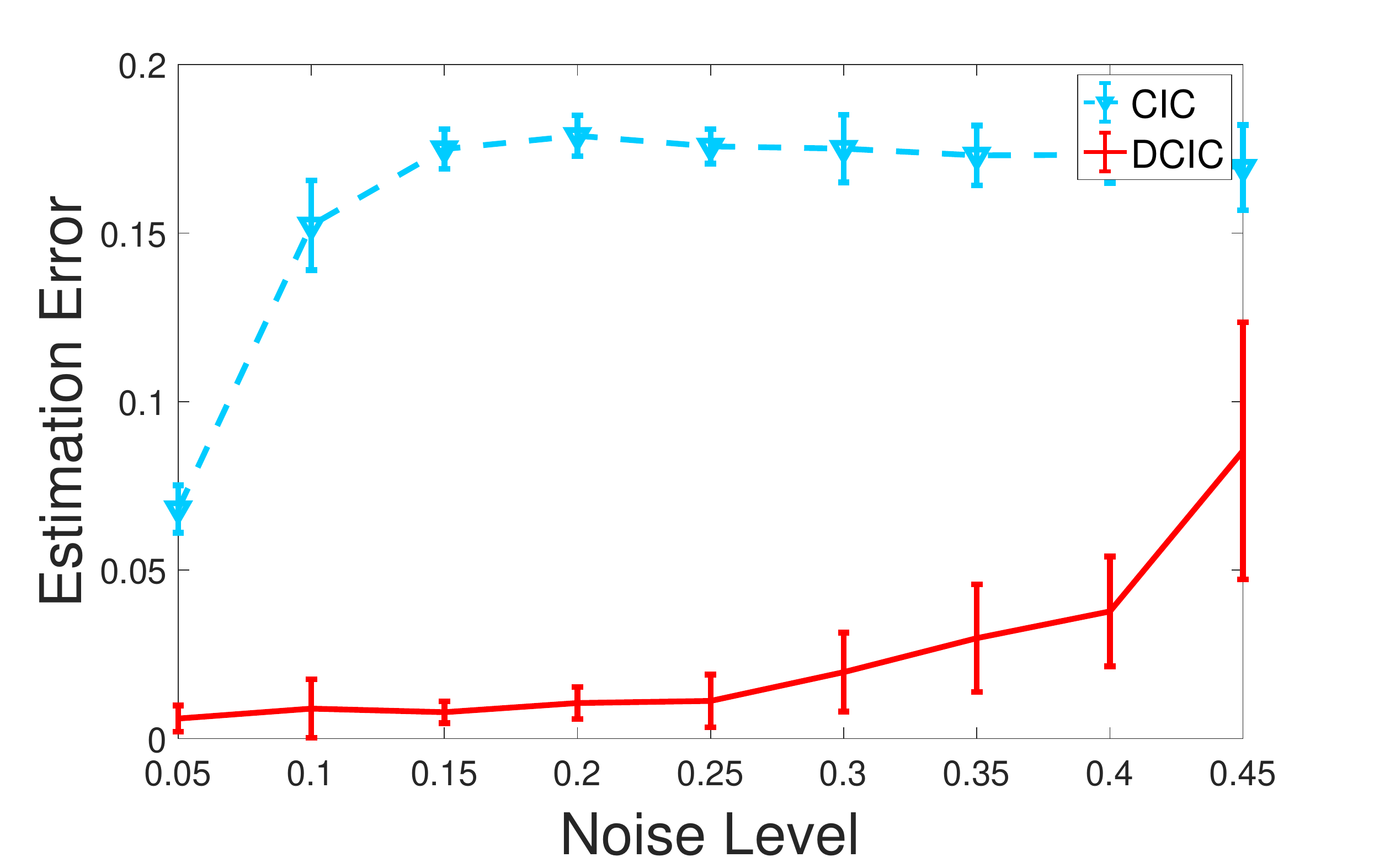}}
\subfigure []
{\hspace{-0pt}\includegraphics[width=0.32\columnwidth]{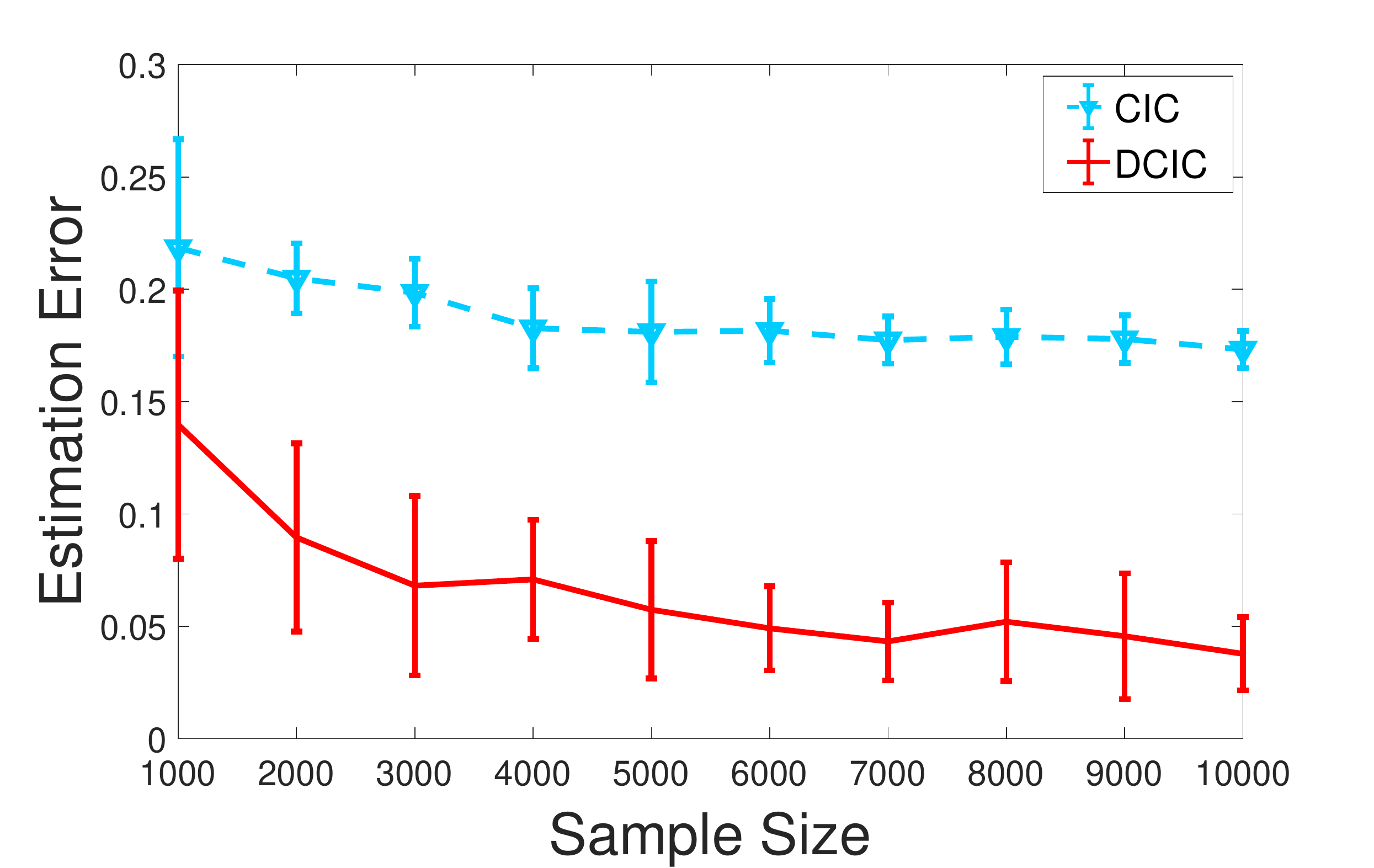}} \label{fig:beta}
\caption{The estimation error of $\beta$. (a), (b), and (c) present the estimate errors with
the increasing class ratio $\beta(Y=1)$, the increasing flip rate $\rho$, and the
increasing sample size $n$, respectively. }
\label{fig:ratio}
\end{figure}

\begin{figure}[t!]
\centering
\subfigure []
{\hspace{-0pt}\includegraphics[width=0.32\columnwidth]{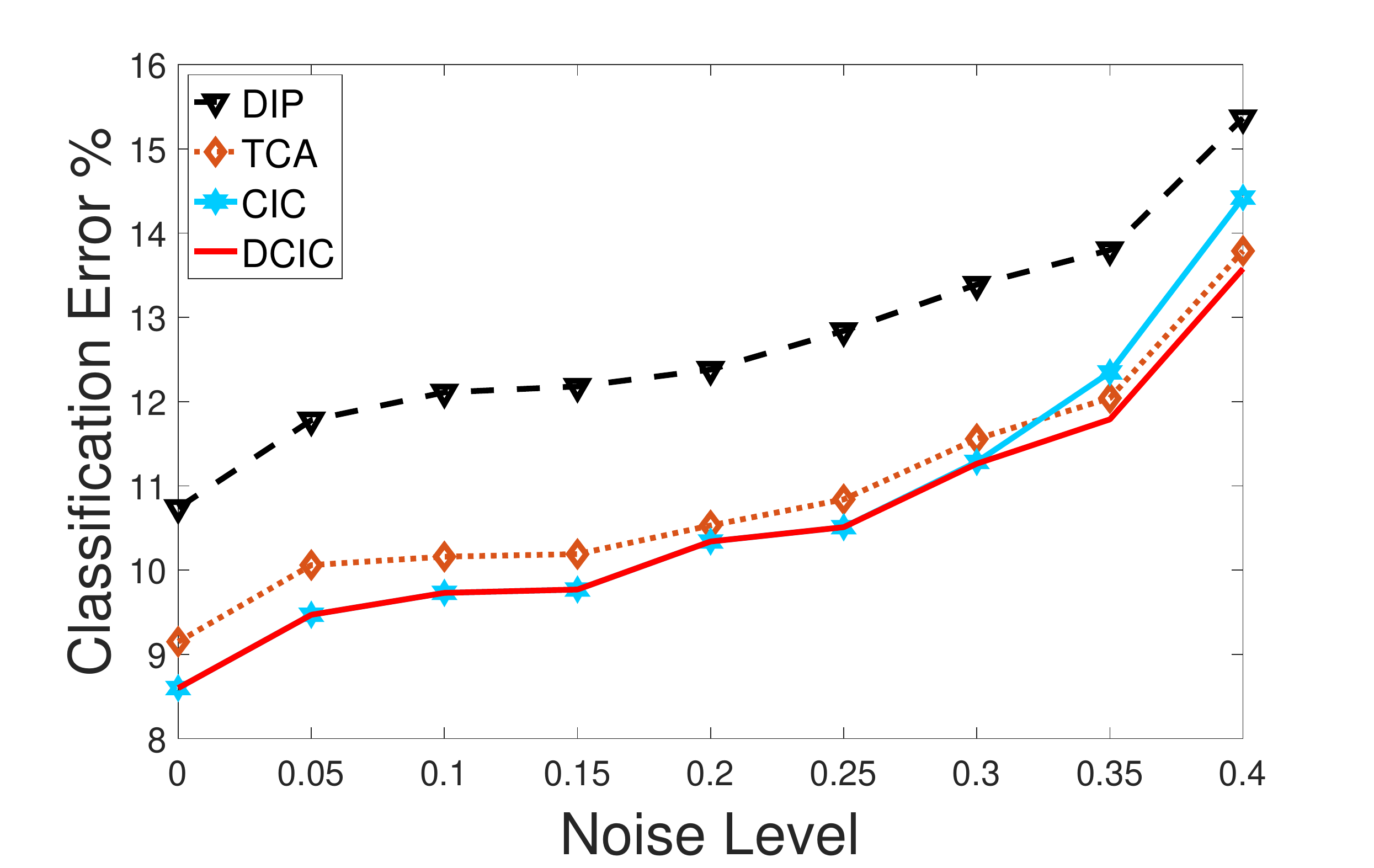}}
\subfigure []
{\hspace{-0pt}\includegraphics[width=0.32\columnwidth]{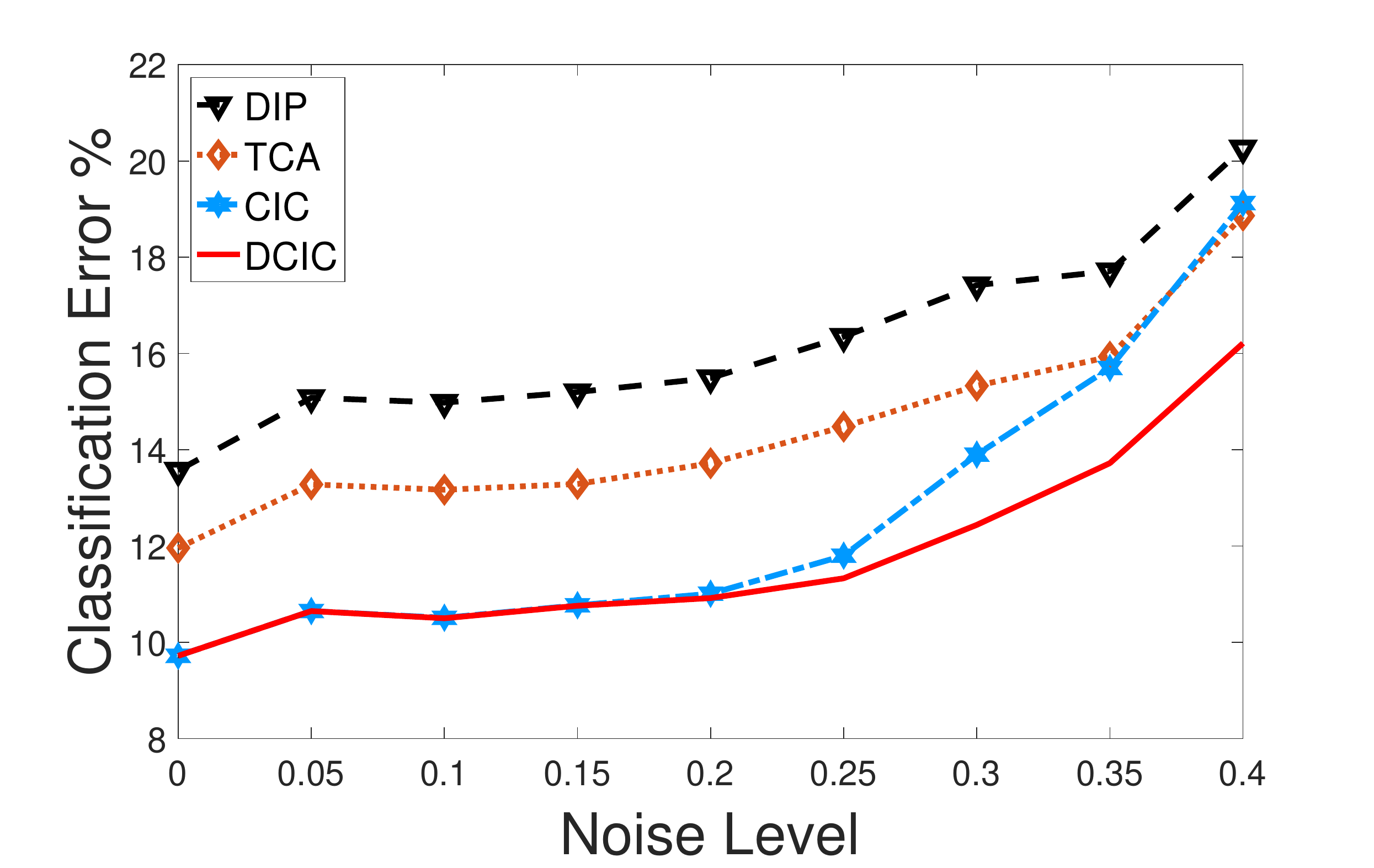}}
\subfigure []
{\hspace{-0pt}\includegraphics[width=0.32\columnwidth]{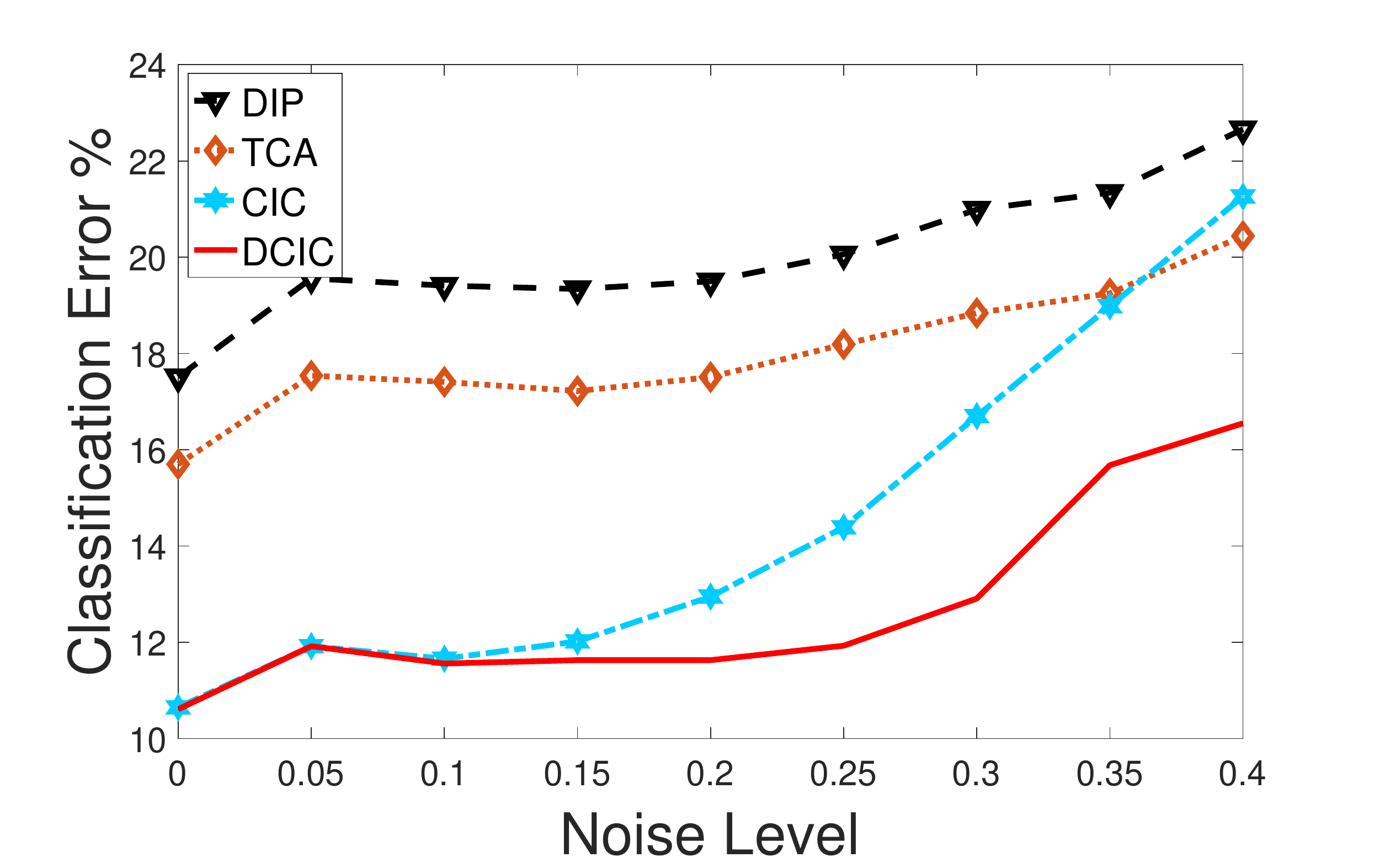}} \label{fig:acc_syn}
\caption{The effectiveness of invariant components extraction. (a), (b), and (c) present
the classification error with increasing flip rate $\rho$ when $\beta_1=1.4$, 1.6, and 1.8, respectively.}
\label{fig:class}
\end{figure}

We study the performance of the linear DCIC model in two situations: (a) the estimation of class ratio $\beta$ in the target shift (TarS) scenario given the true flip rates (i.e., transition probabilities); and (b) the evaluation of the extracted invariant components in the generalized target shift (GeTarS) scenario, with various class ratios and different label flip rates. In all experiments, the flip rates are estimated using the method proposed in \cite{liu2016classification}. We repeat the experiments for 20 times and report the average performances.

We generate the binary classification training and test data from a 2-dimensional mixture of Gaussians \cite{gong2016domain}, i.e., $x\sim \sum_{i=1}^2 \pi_i \mathcal N(\mathbf{\theta}_i, \mathbf{\Sigma}_i)$ where the mean parameters $\mathbf{\theta}_{ij}, j=1,2$ are sampled from the uniform distribution $\mathcal U(-0.25,0.25)$ and the covariance matrices $\mathbf{\Sigma}_i$ are sampled from the Wishart distribution $\mathcal W(2\times \mathbf{I}_2,7)$. The class labels are the cluster indices. Under TarS, $P_{X|Y}$ remains the same. We only change the class priors
across domains. Under GeTarS, we apply location and scale transformations on the features to generate target domain data.
To get the noisy observations, we randomly flip the clean labels in the source domain with
the same transition probability $\rho$.

First, we verify that with corrupted labels, the proposed DCIC can almost recover the correct class ratio under TarS. We set the source class prior $P^{S}(Y=1)$ to 0.5. The target domain class prior $P^{T}(Y=1)$ varies
from 0.1 to 0.9 with step 0.1. The corresponding class ratio $\beta(Y=1)=P^{T}(Y=1)/P^{S}(Y=1)$ varies
from 0.2 to 1.8 with step 0.2. Then, we compare the proposed method with CIC \cite{gong2016domain} on
finding the true class ratio $\beta^*$ with noisy labels in source domain. We evaluate the performance by using the class ratio estimation error $\|\beta_{est}-\beta^*\|/\|\beta^*\|$, where $\beta_{est}$ is the estimated class ratio vector. 
Figure \ref{fig:ratio}(a) shows that DCIC can find the solutions close to the true $\beta^*$ for various class ratios. 
In this experiment, given large label noise ($\rho=0.4$), $\beta$ estimated by CIC is close to the true one only when $\beta^*(Y=1)$ is close to 0, 1, and 2. The estimation of CIC is accurate at $\beta^*(Y=1)=1$ because we set the class prior $P^S_{Y=1}$ to 0.5 in the clean source domain, which happens to make $P_{\rho Y}^S=P_{Y}^S$. If $P^S_{Y=1}\neq 0.5$, then  $P_{\rho Y}^S\neq P_{Y}^S$, the estimated $\beta$ will be wrong (see Section 3). CIC gives accurate results when $\beta^*(Y=1)$ is close to 0, 2 because target domain collapses to a single class, rendering the estimated results trivially right. Figure \ref{fig:ratio}(b) shows the superiority of the proposed method over CIC at different levels of label noise. When $\rho > 0.1$, CIC finds the incorrect solutions. However, our method can find a good solution even when $\rho$ is close to 0.5. Figure \ref{fig:ratio}(c)
shows that the estimate of $\beta$ improves as the sample size gets larger.

Second, under GeTarS, we evaluate whether our method can discover the invariant
representations given the noisy source data and unlabeled target data. In these experiments, we fix the sample size to 500, and the class prior $P^{S}(Y=1)$ to 0.5. We use classification accuracies to measure the performance. The results
in Figure \ref{fig:class} show that our method is more robust to the label noise than DIP, TCA, and CIC.
% This is because the new weights $\beta_{\rho}$ help our method to correct the kernel mean value on the noisy distribution.
\subsection{Real Data}
\begin{table*}[t!]
\small
\centering
\caption{Classification accuracies and their standard deviations for WiFi localization dataset.}
\label{tab:wifi}
\begin{tabular}{l|c|c|c|c|c}
\hline
        & Softmax & TCA  & DIP & CIC & DCIC \\ \hline
t1 $\to$ t2 & 60.73 $\pm$ 0.66 & 70.80 $\pm$ 1.66 & 71.40 $\pm$ 0.83 & 75.50 $\pm$ 1.02 & $\textbf{79.28 $\pm$ 0.56}$               \\ \hline
t1 $\to$ t3 & 55.20 $\pm$ 1.22 & 67.43 $\pm$ 0.55 & 64.65 $\pm$ 0.32 & 69.05 $\pm$ 0.28 & $\textbf{70.75 $\pm$ 0.91}$  \\ \hline
t2 $\to$ t3 & 54.38 $\pm$ 2.01 & 63.58 $\pm$ 1.33 & 66.71 $\pm$ 2.63 & 70.92 $\pm$ 3.86 &  $\textbf{77.28 $\pm$ 2.87}$ \\  \hline
hallway1 & ~~40.81 $\pm$ 12.05 & 42.78 $\pm$ 7.69 & 44.31 $\pm$ 8.34 & 51.83 $\pm$ 8.73 & $~~\textbf{59.31 $\pm$ 12.30}$               \\ \hline
hallway2 & ~~27.98 $\pm$ 10.28 & ~~43.68 $\pm$ 11.07 & 44.61 $\pm$ 5.94 & 43.96 $\pm$ 6.20 & $\textbf{60.50 $\pm$ 8.68}$  \\ \hline
hallway3 & 24.94 $\pm$ 9.89 & 31.44 $\pm$ 5.47 & 33.50 $\pm$ 2.58 & 32.00 $\pm$ 3.88 & $\textbf{33.89 $\pm$ 5.94}$  \\  \hline
\end{tabular}
\end{table*}

\textbf{WiFi Localization Dataset.} We further compare our linear DCIC model with DIP, TCA, and CIC on the cross-domain
indoor WiFi localization dataset \cite{zhang2013covariate}.
% The WiFi localization dataset
% was collected from a hallway area. This hallway area is divided into the spaces of 119
% grids with the size of $1.5 \times 1.5$ square meters each.
The problem is to learn the function
between signals $X$ and locations $Y$. Here, we view it as a
classification problem, where each location space is assigned with a discrete label. In the prediction stage,
the label is then converted to the location information.
We resample the training set to simulate the changes in $P_Y$.
% \textcolor{red}{The first half of the randomly selected classes are assigned with
% a class probability to ensure the class ratio is $2.5$. Then the rest are assigned with another probability
% to ensure the total sum of these probabilities is 1.}
% \textcolor{red}{We randomly sample half of the total classes and let their class ratio be 2.5. The probabilities of the rest
% classes are set to be equal. The class priors in the source domain is thus given, based on which we can resample the training examples.}
To ensure that the class ratio is not a vector of all ones, we resample the source training examples. We randomly select $c/2$ classes and let their class ratios be 2.5. For the other $c/2$ classes, we set their $P(Y)$ to be equal. The flip rate from one label to another is set to $\frac{\rho}{c-1}$.

We first learn the linear transformation $W \in \mathbb{R}^{d\times d'}$ ($d'=10$) and extract the invariant components. A neural network with one hidden layer is trained by minimizing Eq. (\ref{main_framework}) and then obtain the classifier for the signals in target domain according to Eq. (\ref{final_objective}). The output layer is a softmax with the cross-entropy loss. The activation function in the hidden layer is the Rectified Linear Unit (ReLU). The number of neurons in the hidden layer is set to 800. During training, learning rate is fixed to 0.1. After training, as in \cite{gong2016domain}, we report the percentage of examples on which the difference between the predicted and true locations is within 3 meters. Here, we train a neural network with the raw features as the baseline. All the experiments are repeated 10 times and the average performances are reported. In Table \ref{tab:wifi}, the three upper rows present the transfer across different time periods $t1, t2$, and $t3$, where $\rho = 0.4$. The lower part shows the transfer across different devices, where $\rho = 0.2$. We can see that all the results show DCIC can better transfer the invariant knowledge than other methods.

See the results in the lower parts, since the input features in two domains are too complex in these cases, the invariant components cannot be well identified by a simple linear transformation, which finally results in the degraded performances. Therefore, for data with complex features, we would like to introduce our deep denoising models to extract invariant components and to correct the shift. The experiments on deep models are shown in the following subsections.

\begin{table*}[t!]
\small
\centering
\caption{Classification accuracies and their standard deviations for USPS and MNIST datasets.}
\label{tab:uspsmnist}
\begin{tabular}{l|c|c|c|c}
\hline
        &  \begin{tabular}{c} mnist $\to$ usps \\($\rho=0.4$)\end{tabular}  & \begin{tabular}{c} usps $\to$ mnist \\($\rho=0.4$)\end{tabular} & \begin{tabular}{c} mnist $\to$ usps \\($\rho=0.2$)\end{tabular} & \begin{tabular}{c} usps $\to$ mnist \\($\rho=0.2$)\end{tabular} \\ \hline
FT+Forward $Q$ & 58.12 $\pm$ 0.32 & 61.02 $\pm$ 0.90 & 59.27 $\pm$ 1.51 & 65.90 $\pm$ 0.65                \\ \hline
FT+Forward $\hat{Q}$ & 54.93 $\pm$ 2.23 & 60.80 $\pm$ 0.49 & 56.97 $\pm$ 1.36 & 65.51 $\pm$ 3.07                \\ \hline
DAN+Forward $Q$ & 59.34 $\pm$ 5.43 & 64.68 $\pm$ 1.07 & 62.82 $\pm$ 1.15 & 67.05 $\pm$ 0.77  \\ \hline
DAN+Forward $\hat{Q}$ & 54.76 $\pm$ 1.62 & 63.87 $\pm$ 0.84 & 61.28 $\pm$ 1.44 & 65.70 $\pm$ 1.24  \\ \hline
CIC & 65.23 $\pm$ 2.63 & 58.09 $\pm$ 2.17 & 66.70 $\pm$ 1.31 & 61.02 $\pm$ 3.96                \\ \hline
CIC+Forward $Q$ & 65.37 $\pm$ 2.49 & 63.35 $\pm$ 4.43 & 66.84 $\pm$ 3.62 & 68.45 $\pm$ 0.91  \\ \hline
CIC+Forward $\hat{Q}$ & 64.18 $\pm$ 1.49 & 62.78 $\pm$ 2.92 & 63.42 $\pm$ 0.99 & 67.99 $\pm$ 1.30  \\ \hline
DCIC+Forward $Q$ & $\textbf{69.94 $\pm$ 2.25}$ & $\textbf{68.77 $\pm$ 2.34}$ & $\textbf{72.33 $\pm$ 2.15}$ & $\textbf{70.80 $\pm$ 1.59}$  \\ \hline
DCIC+Forward $\hat{Q}$ & $\textbf{68.50 $\pm$ 0.37}$ & $\textbf{66.78 $\pm$ 1.53}$ & $\textbf{69.29 $\pm$ 4.07}$ & $\textbf{70.47 $\pm$ 2.29}$  \\ \hline
% Upper Bound & 75.04 & 73.36 & 75.04 & 73.36  \\ \hline
\end{tabular}
\end{table*}

\textbf{MNIST-USPS.} USPS dataset is a handwritten digit dataset including ten classes 0-9 and contains 7,291 training images and 2,007 test images of size $16\times 16$, which is rescaled to $28 \times 28$. MNIST shares the same 10 classes of digits which consist of 60,000 training images and 10,000 test images of size $28\times 28$. In our experiments, these two datasets are resampled to construct the transfer learning datasets in which the class priors $P_Y$ across different domains vary. For MNIST, we assume that the class priors are unbalanced. For the first 5 classes, the class prior is set to 0.04. For the rest 5 classes, the class prior is equal to 0.16. For USPS, the class priors are balanced; that is, the class prior is set to 0.1 for each class. According to these class priors, we sample 5,000 images from both MNIST and USPS datasets to construct the new dataset mnist2usps. We switch the source/target pair to get another dataset usps2mnist. Same with \cite{patrini2017making}, in the source data, noise flips between the similar digits: $2\to 7$, $3\to 8$, $5\leftrightarrow 6$, $7\to 1$ with the transition probability $\rho=0.2$ or $0.4$. After the noisy data are obtained, we leave 10 percent of source data as validation set. The LeNet \cite{lecun1998gradient} structure in Caffe's \cite{jia2014caffe} MNIST tutorial is employed to train the model from the scratch. Our denoising MMD loss is imposed on the first fully connected layer. In all experiments, $l_2$ regularization is applied and we set $\pi_1=1$ and $\pi_2=1e-4$. The batch sizes for both source and target data are set to 100. The initial learning rate $r_0=0.01$ and is decayed exponentially according to $r_0(1+0.0001t)^{-0.75}$, where $t$ is the index of current iteration. Each experiment is repeated 5 times. % Other experimental details are presented in the Supplementary Material.

%The LeNet structure used in the experiments on MNIST+USPS datasets are $conv1 (5,5,1,20) \to pooling (2,2) \to conv2 (5,5,20,50) \to pooling (2,2) \to fc1 (800, 500) \to fc2 (500,10)$, where $conv (h\times w\times c_i \times c_o)$ represents the convoultional layer with $c_i$ input channels, $c_o$ output channels, and kernel with size of $h \times w$; and fc ($d_i \times d_o$) represents the fully connected layer whose weight is a $d_i \times d_o$ matrix. The ReLU activation function is used after the $fc1$ layer. The proposed denoising MMD loss is also imposed on $fc1$ layer. 

Here, DCIC is compared with the baseline that finetunes the source data only (FT), DAN, and CIC. These methods are integrated with the forward procedure in \cite{patrini2017making} to reduce the effects of label noise. They are denoted as methods with ``Forward $Q$ (resp. $\hat{Q}$)'' given the true (resp. estimated) transition matrix. The results are shown in Table \ref{tab:uspsmnist}. When label noise is present, CIC based methods cannot correctly estimate the class ratios, which adversely affects the identification of the invariant components. It thus performs worse than the DAN based methods in some cases. The latter, however, ignores the change of $P_Y$ in different domains. In contrast, our method often gives better estimation of the class ratios and can effectively identify the invariant components, which leads to the higher performances. 

\textbf{VLCS.} VLCS dataset \cite{torralba2011unbiased} consists of the images from five common classes: ``bird'', ``car'', ``chair'', ``dog'', and ``person'' in the datasets Pascal VOC 2007 (V), LabelMe (L), Caltech (C), and SUN09 (S), respectively. For these four datasets, we first randomly select at most 300 images for each class to construct the new datasets, respectively. Then, we construct the transfer learning datasets by using the leave-one-domain-out evaluation strategy. For example, in ``VLS2C'', the source data is the combination of the new Pascal VOC 2007, LabelMe, and SUN09 datasets. The target dataset is the new Caltech. In each source data, the labels flip from ``person'' to ``car'', ``chair'' to ``person'', and ``dog'' to ``person'' with the probability $\rho=0.4$. We leave $30\%$ of the source data as the validation set. Each experiment is repeated 5 times.

In this experiments, the source data is finetuned on the pretrained AlexNet \cite{krizhevsky2012imagenet} model with the parameters in conv1-conv3 layers being freezed. We impose our denoising MMD loss on the fc7 layer. The batch sizes for both source and target data are 32. The initial learning rate is 0.001 and decayed exponentially according to $0.001(1+0.002t)^{-0.75}$. The results are shown in Table \ref{tab:vlcs}. Our proposed method also improves the performances of the compared baselines, which indicates the effectiveness of the proposed model to correct the shift in different domains even though the label noise is present.

\begin{table*}[t!]
\small
\centering
\caption{Classification accuracies and their standard deviations for VLCS dataset.}
\label{tab:vlcs}
\begin{tabular}{l|c|c|c|c}
\hline
        &  VLS2C & LCS2V & VLC2S & VCS2L \\ \hline
FT+Forward $Q$ & 85.88 $\pm$ 2.17 & 62.07 $\pm$ 0.86 & 59.40 $\pm$ 1.37 & 49.34 $\pm$ 1.39       \\ \hline
FT+Forward $\hat{Q}$ & 78.62 $\pm$ 4.36 & 59.49 $\pm$ 0.50 & 57.09 $\pm$ 1.81 & 49.14 $\pm$ 1.39   \\ \hline
DAN+Forward $Q$ & 87.66 $\pm$ 2.37 & 64.37 $\pm$ 2.07 & 59.54 $\pm$ 0.83 & 51.07 $\pm$ 1.26  \\ \hline
DAN+Forward $\hat{Q}$ & 84.69 $\pm$ 0.24 & 58.64 $\pm$ 1.91 & 57.51 $\pm$ 1.25 & 50.41 $\pm$ 1.20  \\ \hline
CIC & 75.15 $\pm$ 6.23 & 54.69 $\pm$ 0.96 & 53.61 $\pm$ 2.35 & 49.30 $\pm$ 0.48            \\ \hline
CIC+Forward $Q$ & 86.83 $\pm$ 2.53 & 64.22 $\pm$ 0.27 & 60.36 $\pm$ 0.36 & 51.76 $\pm$ 0.82  \\ \hline
CIC+Forward $\hat{Q}$ & 85.69 $\pm$ 1.76 & 59.80 $\pm$ 0.47 & 57.65 $\pm$ 0.60 & 50.33 $\pm$ 0.31  \\ \hline
DCIC+Forward $Q$ & $\textbf{91.60 $\pm$ 0.51}$ & $\textbf{65.67 $\pm$ 0.37}$ & $\textbf{61.79 $\pm$ 0.77}$ & $\textbf{52.47 $\pm$ 0.50}$  \\ \hline
DCIC+Forward $\hat{Q}$ & 87.28 $\pm$ 1.18 & 63.35 $\pm$ 0.37 & 58.88 $\pm$ 0.74 & 51.60 $\pm$ 1.48  \\ \hline
% Upper Bound & 75.04 & 73.36 & 75.04 & 73.36  \\ \hline
\end{tabular}
\end{table*}

%\begin{table}[h]
%\small
%\vspace{-5pt}
%\centering
%\caption{Classification accuracies and their standard deviations for different transfer learning methods in the presence of label noise.}
%\label{tab:wifi}
%\begin{tabular}{l|c|c|c|c|c}
%\hline
%        &  FT/F  & DAN/F & CIC & CIC/F & DCIC/F \\ \hline
%\begin{tabular}{c} mnist $\to$ usps \\($\rho=0.4$)\end{tabular} & 54.76 $\pm$ 2.23 & 54.93 $\pm$ 1.62 & 65.23 $\pm$ 2.63 & 65.37 $\pm$ 2.49 & $\textbf{68.50 $\pm$ 0.37}$               \\ \hline
%\begin{tabular}{c} usps $\to$ mnist \\($\rho=0.4$)\end{tabular} & 61.02 $\pm$ 0.90 & 63.87 $\pm$ 0.84 & 58.09 $\pm$ 2.17 & 63.35 $\pm$ 4.43 & $\textbf{66.78 $\pm$ 1.53}$  \\ \hline
%\begin{tabular}{c} mnist $\to$ usps \\($\rho=0.2$)\end{tabular} & 60.73 $\pm$ 0.66 & 70.80 $\pm$ 1.66 & 71.40 $\pm$ 0.83 & 75.50 $\pm$ 1.02 & $\textbf{79.28 $\pm$ 0.56}$               \\ \hline
%\begin{tabular}{c} usps $\to$ mnist \\($\rho=0.2$)\end{tabular} & 55.20 $\pm$ 1.22 & 67.43 $\pm$ 0.55 & 64.65 $\pm$ 0.32 & 69.05 $\pm$ 0.28 & $\textbf{70.75 $\pm$ 0.91}$  \\ \hline
%\end{tabular}
%\vspace{-0pt}
%\end{table}

\subsection{Discussions}
\subsubsection{Convergence analysis}
In order to verify the effectiveness of the proposed method to estimate $P^T_Y$, in Figure \ref{fig_estimation_error} (a), we show the convergence of the estimation errors $\frac{\|\alpha^*-\hat{\alpha}\|_2}{\|\alpha^*\|_2}$ of our ``DCIC + Forward $\hat{Q}$'' method and the ``CIC + Forward $\hat{Q}$'' method, where $\alpha^*$ is the true class prior and $\hat{\alpha}$ is the estimated one. The experiment is conducted on the mnist2usps dataset. We can see that our proposed method can find a better solution for $P^T_Y$ after using our denoising MMD loss.

\subsubsection{Parameter sensitivity}
\begin{figure}[t!]
\centering
\subfigure []
{\hspace{-0pt}\includegraphics[width=0.48\columnwidth]{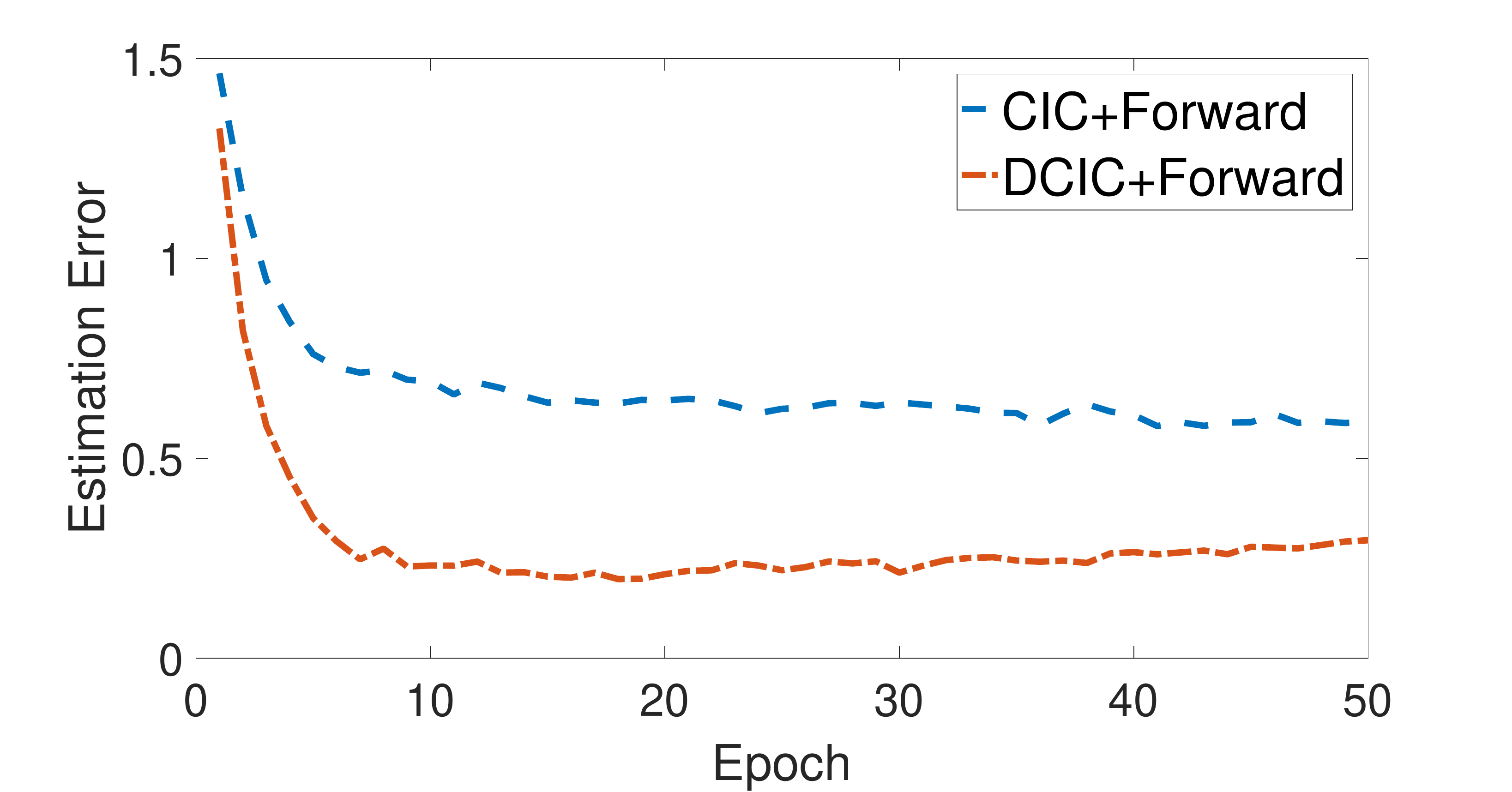}}
\subfigure []
{\hspace{-0pt}\includegraphics[width=0.48\columnwidth]{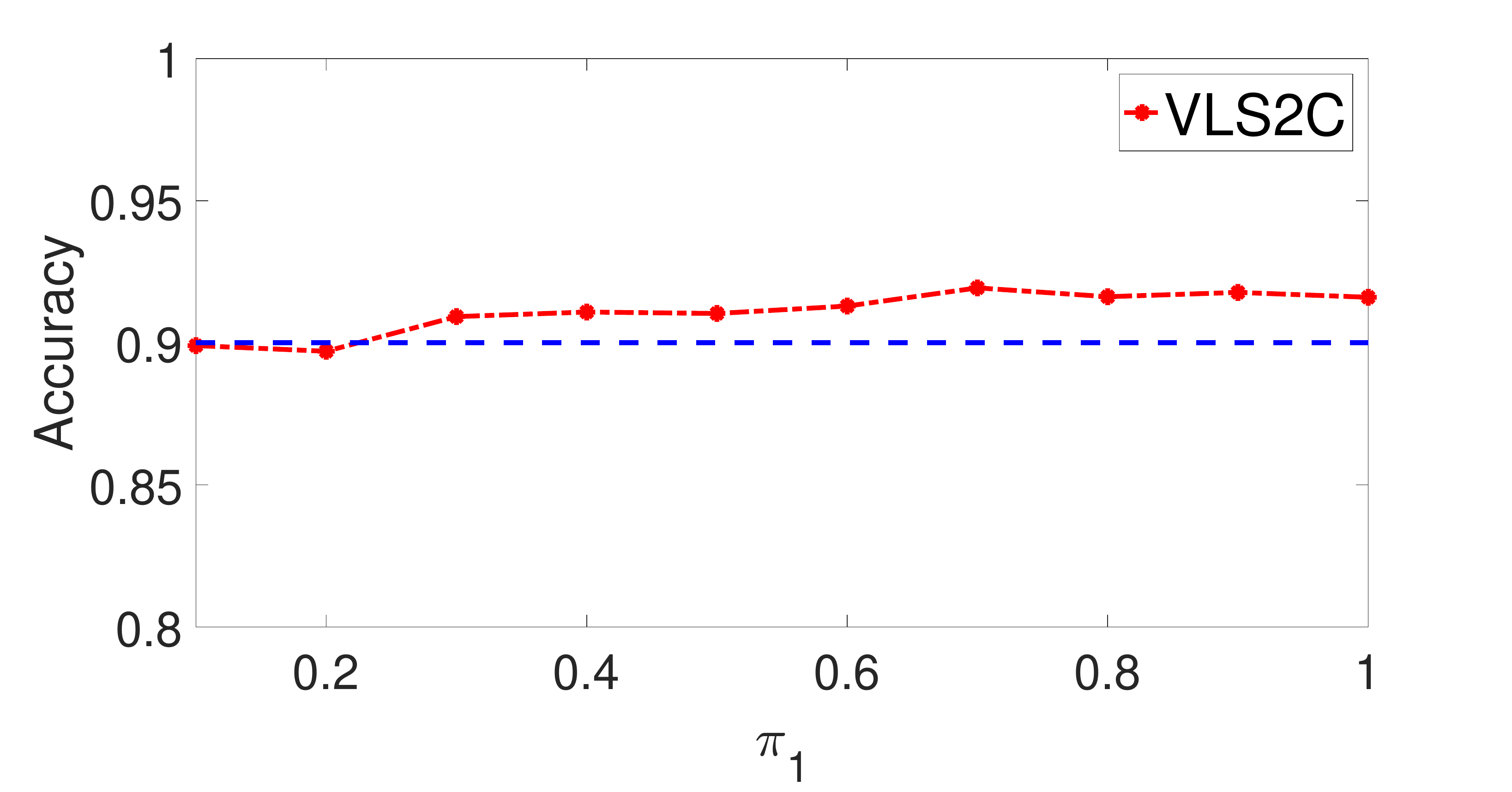}}
\caption{(a) The convergence of class prior estimation in target domain. (b) The sensitivity analysis of the parameter $\pi_1$.}
\label{fig_estimation_error}
\end{figure}

Here, we check the sensitivity of the trade-off parameter $\pi_1$ of our denoising MMD loss. Figure \ref{fig_estimation_error} (b) shows the classification accuracies with respect to different values of $\pi_1$, which ranges from $0.1$ to $1.0$ with step $0.1$. This task is evaluated on VLS2C dataset. We can see, the overall performance is not very sensitive to the choice of $\pi_1$. In our experiments, we find $\pi_1=1.0$ works well on all other datasets.

\section{Conclusion}
In this paper, we have studied the problem of transfer learning with label noise. We have found that the presence of labels is detrimental to the performance of existing transfer learning methods. In particular, when the label is the cause for the features, the estimate of target domain class distribution and conditional invariant representations can be unreliable. To alleviate the effects of label noise on transfer learning, we have proposed the new transfer learning models which employs the novel denoising MMD loss to improve the estimation of both target domain label distribution and conditional invariant components from the noisy source data and the unlabeled target data. We have provided both theoretical and empirical studies to demonstrate the effectiveness of the proposed method. 

\newpage

\appendix

% \maketitle

\section{Proof of Theorem 1} \label{appA}
\label{app:theorem}

% Note: in this sample, the section number is hard-coded in. Following
% proper LaTeX conventions, it should properly be coded as a reference:

%In this appendix we prove the following theorem from
%Section~\ref{sec:textree-generalization}:

$\textbf{Proof.}$
In this proof, $Y=y$ (resp. $\hat{Y} = y'$) is replaced by $y$ (resp. $y'$) for simplicity. For example, we let $P^S(\hat{Y}=y'|Y=y)=P^S(y'|y)$. We also let $X'=\tau(X)$. According to Eq. (2), we have
\begin{equation}
P^{\textrm{new}}_{X'} = \sum_y \sum_{y'} \beta_{\rho}(y') P^S(y'|X',y) P^S(X',y) = \sum_y P^S(X'|y) P^S(y) \sum_{y'} P^S(y'|y) \beta_{\rho}(y'). 
\end{equation}

Because $P^T_{X'} = \sum_y P^T(X'|y) P^T(y)$, then combining with the above equation, we have
\begin{equation}
\sum_y P^T(X'|y) P^T(y) = \sum_y P^S(X'|y) P^S(y)\sum_{y'} P^S(y'|y) \beta_{\rho}(y').
\end{equation}

Because the transformation $\tau$ satisfies that $P(X'|Y=i), i \in \{1,\cdots,c\}$ are linearly independent, there exist $\textbf{no}$ such non-zero $\gamma_1,\cdots,\gamma_c$ and $\kappa_1,\cdots,\kappa_c$ that  $\sum_{i=1}^c \gamma_i P^S(X'|Y=i) = 0$ and $\sum_{i=1}^c \kappa_i P^T(X'|Y=i)=0$. According to the assumption in Theorem 1, the elements in the set $\{v_i P^S(X'|Y=i) + \lambda_i P^T(X'|Y=i); i \in \{1,\cdots,c\}; \forall v_i, \lambda_i \textrm{ } (v_i^2 + \lambda_i^2 \neq 0)\}$ are also linearly independent. Then we have, $\forall y \in \{1,\cdots,c\}$,
\begin{equation} 
P^T(X'|y) P^T(y) - P^S(X'|y) P^S(y) \sum_{y'} P^S(y'|y) \beta_{\rho}(y') = 0.
\end{equation}

Taking the integral of above equation w.r.t. $X'$, we have
\begin{equation} \label{eq:beta_rela}
P^T(y) = P^S(y) \sum_{y'} P^S(y'|y) \beta_{\rho}(y'),
\end{equation}
which further implies $P^T(X'|y) = P^S(X'|y), \forall y \in \{1,\cdots,c\}$.
According to Eq. (\ref{eq:beta_rela}), we have $\forall y \in \{1,\cdots,c\},$
\[\sum_{y'} P^S(y'|y) \beta_{\rho}(y') = P^T(y)/P^S(y) = \beta(y).\]
The proof of Theorem 1 ends. \hfill$\blacksquare$

\section{Proof of Theorem 2} \label{appB}
Recall the denoising MMD loss, we have
\[\hat{\mathcal{D}}(W,\alpha) = \|\frac{1}{m}\psi(\mathbf{x'}^{S})G\alpha- \frac{1}{n}\psi(\mathbf{x'}^{T})\mathbf{1}\|^2.\]
Let
\[\mathcal{D}(W,\alpha)=\|\mathbb{E}\frac{1}{m}\psi(\mathbf{x'}^{S})G\alpha-\mathbb{E} \frac{1}{n}\psi(\mathbf{x'}^{T})\mathbf{1}\|^2,\]
where we abuse the training samples $\{(x_1^S,\hat{y}_1^S),\cdots,(x_m^S,\hat{y}_m^S)\}$ and $\{x_1^T,\cdots,x_n^T\}$  as being i.i.d. variables, respectively.

We analyze the convergence property of the learned $\hat{\alpha}$ to the optimal one $\alpha^*$ by analyzing the convergence from the expected objective function $\mathcal{D}(\hat{W},\hat{\alpha})$ to $\mathcal{D}(\hat{W},\alpha^*)$.

To prove Theorem 2, we need the following Theorem \ref{micdiarmid}, Lemma \ref{le1}, and Lemma \ref{le2}. Theorem \ref{micdiarmid} is about concentration inequality (McDiarmid's inequality \cite{boucheron2013concentration}, also known as the bounded difference inequality). Lemma \ref{le1} shows that the distance $\mathcal{D}(\hat{W},\hat{\alpha})-\mathcal{D}(\hat{W},\alpha^*)$ can be upper bounded even though we do not know the optimal $\alpha^*$. Lemma \ref{le2} upper bounds the Rademacher-like \cite{bartlett2002rademacher} term $\mathbb{E}\sup_{\alpha\in \Delta}\|f(\mathbf{x}^{S},\mathbf{x}^{T},\alpha)\|^2$.

\begin{theorem}\label{micdiarmid}
Let $X=[X_1,\cdots,X_n]$ be an independent and identically distributed sample and $X^i$ a new sample with the $i$-th example in $X$ being replaced by an independent example $X'_i$. If there exists $b_1,\cdots,b_n>0$ such that $f:\mathcal{X}^n\rightarrow \mathbb{R}$ satisfies the following conditions
\[|f(X)-f(X^i)|\leq b_i, \forall i\in\{1,\cdots,n\}.\]
Then for any $X\in\mathcal{X}^n$ and $\epsilon>0$, the following inequality holds
\[P(\mathbb{E}f(X)-f(X)\geq\epsilon)\leq\exp\left(\frac{-2\epsilon^2}{\sum_{i=1}^{n}b_i^2}\right).\]
\end{theorem}

\begin{lemma}\label{le1}
We denote $  \Delta\triangleq\{\alpha|\alpha\geq 0, \|\alpha\|_1=1\}$ and
\begin{equation}
f(\mathbf{x}^{S},\mathbf{x}^{T},\alpha) \triangleq \mathbb{E}\left(\frac{1}{m}\psi(\mathbf{x'}^{S})G\alpha-\frac{1}{n}\psi(\mathbf{x'}^{T})\mathbf{1}\right) -\frac{1}{m}\psi(\mathbf{x'}^{S})G\alpha+\frac{1}{n}\psi(\mathbf{x'}^{T})\mathbf{1}.
\end{equation}
Then, we have
\begin{equation}
\begin{aligned}
\mathcal{D}(\hat{W},\hat{\alpha})-\mathcal{D}(\hat{W},\alpha^*) &\leq 2\sup_{\alpha\in \Delta}|\mathcal{D}(\hat{W},{\alpha})-\hat{\mathcal{D}}(\hat{W},\alpha)| \\
&\leq 4(\wedge_{\hat{Q}}+1)\wedge_{\hat{W}}\sup_{\alpha\in \Delta}\|f(\mathbf{x}^{S},\mathbf{x}^{T},\alpha)\|.
\end{aligned}
\end{equation}
\end{lemma}

$\textbf{Proof.}$
We have
\begin{equation}
\begin{aligned}
&\mathcal{D}(\hat{W},\hat{\alpha})-\mathcal{D}(\hat{W},\alpha^*)\\
&=\mathcal{D}(\hat{W},\hat{\alpha})-\hat{\mathcal{D}}(\hat{W},\hat{\alpha}) +\hat{\mathcal{D}}(\hat{W},\hat{\alpha})- \hat{\mathcal{D}}(\hat{W},\alpha^*)+\hat{\mathcal{D}}(\hat{W},\alpha^*)-\mathcal{D}(\hat{W},\alpha^*)\\
&\leq \mathcal{D}(\hat{W},\hat{\alpha})-\hat{\mathcal{D}}(\hat{W},\hat{\alpha})+\hat{\mathcal{D}}(\hat{W},\alpha^*)-\mathcal{D}(\hat{W},\alpha^*)\\
&\leq 2\sup_{\alpha\in \Delta}|\mathcal{D}(\hat{W},{\alpha})-\hat{\mathcal{D}}(\hat{W},\alpha)|,
\end{aligned}
\end{equation}
where the first inequality holds because $\hat{\alpha}$ is the empirical minimizer of $\hat{\mathcal{D}}(\hat{W},{\alpha})$ and thus $\hat{\mathcal{D}}(\hat{W},\hat{\alpha})\leq \hat{\mathcal{D}}(\hat{W},{\alpha^*})$.

Further, we have
\begin{equation}
\begin{aligned}
&|\mathcal{D}(\hat{W},{\alpha})-\hat{\mathcal{D}}(\hat{W},\alpha)|\\
&=\left(\mathbb{E}\left(\frac{1}{m}\psi(\mathbf{x'}^{S})G\alpha-
\frac{1}{n}\psi(\mathbf{x'}^{T})\mathbf{1}\right)+\frac{1}{m}\psi(\mathbf{x'}^{S})G\alpha-\frac{1}{n}\psi(\mathbf{x'}^{T})\mathbf{1}\right)^\top\\
&\left(\mathbb{E}\left(\frac{1}{m}\psi(\mathbf{x'}^{S})G\alpha-
\frac{1}{n}\psi(\mathbf{x'}^{T})\mathbf{1}\right)-\frac{1}{m}\psi(\mathbf{x'}^{S})G\alpha+
\frac{1}{n}\psi(\mathbf{x'}^{T})\mathbf{1}\right)\\
&\leq \left\| \mathbb{E}\left(\frac{1}{m}\psi(\mathbf{x'}^{S})G\alpha-
\frac{1}{n}\psi(\mathbf{x'}^{T})\mathbf{1}\right)+\frac{1}{m}\psi(\mathbf{x'}^{S})G\alpha-\frac{1}{n}\psi(\mathbf{x'}^{T})\mathbf{1} \right\|\\
& \left\|\mathbb{E}\left(\frac{1}{m}\psi(\mathbf{x'}^{S})G\alpha-
\frac{1}{n}\psi(\mathbf{x'}^{T})\mathbf{1}\right)-\frac{1}{m}\psi(\mathbf{x'}^{S})G\alpha+
\frac{1}{n}\psi(\mathbf{x'}^{T})\mathbf{1}\right\|\\
&\leq 2(\wedge_{\hat{Q}}+1)\wedge_{\hat{W}} \left\|\mathbb{E}\left(\frac{1}{m}\psi(\mathbf{x'}^{S})G\alpha-
\frac{1}{n}\psi(\mathbf{x'}^{T})\mathbf{1}\right)-\frac{1}{m}\psi(\mathbf{x'}^{S})G\alpha+
\frac{1}{n}\psi(\mathbf{x'}^{T})\mathbf{1}\right\|,
\end{aligned}
\end{equation}
where the first inequality holds because of Cauchy-Schwarz inequality.

Since
\begin{equation}
f(\mathbf{x}^{S},\mathbf{x}^{T},\alpha) \triangleq \mathbb{E}\left(\frac{1}{m}\psi(\mathbf{x'}^{S})G\alpha-
\frac{1}{n}\psi(\mathbf{x'}^{T})\mathbf{1}\right)-\frac{1}{m}\psi(\mathbf{x'}^{S})G\alpha+\frac{1}{n}\psi(\mathbf{x'}^{T})\mathbf{1},
\end{equation}
we have
\begin{equation}
2\sup_{\alpha\in \Delta}|\mathcal{D}(\hat{W},{\alpha})-\hat{\mathcal{D}}(\hat{W},\alpha)| \leq 4(\wedge_{\hat{Q}}+1)\wedge_{\hat{W}}\sup_{\alpha\in \Delta}\|f(\mathbf{x}^{S},\mathbf{x}^{T},\alpha)\|.
\end{equation}
The proof ends. \hfill$\blacksquare$

\begin{lemma}\label{le2}
Given learned $\hat{Q}$ and $\hat{W}$, let the induced RKHS be universal and upper bounded that $\|\psi(\tau(x))\|\leq\wedge_{\hat{W}}$ for all $x$ in the source and target domains. Let the entries of $G$ be bounded that $|G_{ij}|\leq\wedge_{\hat{Q}}$ for all $i\in \{1,\cdots,m\}, j\in\{1,\cdots,c\}$. We have
\begin{eqnarray*}
\mathbb{E}\sup_{\alpha\in \Delta}\|f(\mathbf{x}^{S},\mathbf{x}^{T},\alpha)\|^2\leq4(\wedge_{\hat{Q}}+1)^2\wedge_{\hat{W}}^2\sqrt{c}(\frac{1}{\sqrt{m}}+\frac{1}{\sqrt{n}}).
\end{eqnarray*}
\end{lemma}

$\textbf{Proof.}$
Recall that when  $\hat{y}_k=i$, $\forall k \in \{1,\cdots,m\}$, the $k$-th row of $G\in \mathbb{R}^{m\times c}$ is $[\frac{\hat{Q}^{-1}_{i1}}{\hat{P}^{S}(Y=1)},\cdots,\frac{\hat{Q}^{-1}_{ic}}{\hat{P}^{S}(Y=c)}]$.
Given $\hat{Q}$, $\hat{W}$ and the estimated $\hat{P}^S(Y)$, we assumed that the entries of $G$ is bounded, i.e., $|G_{ij}|\leq\wedge_{\hat{Q}}$, and that RKHS is upper bounded, i.e., $-\psi_{\max}\leq\psi(\tau(x))\leq\psi_{\max}$ and $\|\psi_{\max}\|\leq\wedge_{\hat{W}}$. Because $\alpha\geq 0$ and $\|\alpha\|_1=1$, we can conclude that for any training sample in the source domain, we have
\[\|\frac{1}{m}\psi(\mathbf{x'}^{S})G\alpha\|\leq\wedge_{\hat{W}}\wedge_{\hat{Q}}.\]
We then have $\|f(\mathbf{x}^{S},\mathbf{x}^{T},\alpha)\|\leq2(\wedge_{\hat{Q}}+1)\wedge_{\hat{W}}$ and that \[\|f(\mathbf{x}^{S},\mathbf{x}^{T},\alpha)\|^2\leq 2(\wedge_{\hat{Q}}+1)\wedge_{\hat{W}}\|f(\mathbf{x}^{S},\mathbf{x}^{T},\alpha)\|.\]
Accordingly, we have
\begin{equation}\label{rade1}
\mathbb{E}\sup_{\alpha\in \Delta}\|f(\mathbf{x}^{S},\mathbf{x}^{T},\alpha)\|^2 \leq2(\wedge_{\hat{Q}}+1)\wedge_{\hat{W}}\mathbb{E}\sup_{\alpha\in \Delta}\|f(\mathbf{x}^{S},\mathbf{x}^{T},\alpha)\|.
\end{equation}
Furthermore, let $\tilde{\mathbf{x}}^{S}$ and $\tilde{\mathbf{x}}^{T}$ be i.i.d. copies of $\mathbf{x}^{S}$ and $\mathbf{x}^{T}$, respectively. In the literature, $\tilde{\mathbf{x}}^{S}$ and $\tilde{\mathbf{x}}^{T}$ are referred as ghost samples \cite{mohri2012foundations}. We have
\begin{equation}\label{rade2}
\begin{aligned}
&\mathbb{E}\sup_{\alpha\in \Delta}\|f(\mathbf{x}^{S},\mathbf{x}^{T},\alpha)\|\nonumber\\
&=\mathbb{E}\sup_{\alpha\in \Delta}\left\|\mathbb{E}\left(\frac{1}{m}\psi(\mathbf{x'}^{S})G\alpha-\frac{1}{n}\psi(\mathbf{x'}^{T})\mathbf{1}\right)  -\frac{1}{m}\psi(\mathbf{x'}^{S})G\alpha+\frac{1}{n}\psi(\mathbf{x'}^{T})\mathbf{1}\right\|\nonumber\\
&=\mathbb{E}_{\mathbf{x}^{S},\mathbf{x}^{T}}\sup_{\alpha\in \Delta}\left\|\mathbb{E}_{\tilde{\mathbf{x}}^{S},\tilde{\mathbf{x}}^{T}}\left(\frac{1}{m}\psi(\tilde{\mathbf{x}}'^{S})G\alpha-\frac{1}{n}\psi(\tilde{\mathbf{x}}'^{T})\mathbf{1}\right) -\frac{1}{m}\psi(\mathbf{x'}^{S})G\alpha+
\frac{1}{n}\psi(\mathbf{x'}^{T})\mathbf{1}\right\|\nonumber\\
&\leq\mathbb{E}_{\mathbf{x}^{S},\mathbf{x}^{T},\tilde{\mathbf{x}}^{S},\tilde{\mathbf{x}}^{T}}\sup_{\alpha\in \Delta}\left\|\left(\frac{1}{m}\psi(\tilde{\mathbf{x}}'^{S})G\alpha-\frac{1}{n}\psi(\tilde{\mathbf{x}}'^{T})\mathbf{1}\right) -\frac{1}{m}\psi(\mathbf{x'}^{S})G\alpha+
\frac{1}{n}\psi(\mathbf{x'}^{T})\mathbf{1}\right\|,
\end{aligned}
\end{equation}
where the last inequality holds because of Jensen's inequality and that every norm is a convex function.

Since $\tilde{\mathbf{x}}^{S}$ and $\tilde{\mathbf{x}}^{T}$ be i.i.d. copies of $\mathbf{x}^{S}$ and $\mathbf{x}^{T}$, respectively, the random variable $\frac{1}{m}\psi(\tilde{\mathbf{x}}'^{S})G\alpha-\frac{1}{n}\psi(\tilde{\mathbf{x}}'^{T})\mathbf{1}-\frac{1}{m}\psi(\mathbf{x'}^{S})G\alpha+\frac{1}{n}\psi(\mathbf{x'}^{T})\mathbf{1}$ is a symmetric random variable, which means its density function is even.
Let $\sigma_i$ be independent Rademacher variables, which are uniformly distributed from $\{-1,1\}$. Let
\[\psi(\mathbf{x'}^{S},\sigma)\triangleq[\sigma_1\psi({x}_1'^{S}),\cdots,\sigma_m\psi({x}_m'^{S})]^\top;\] and
\[\psi(\mathbf{x'}^{T},\sigma)\triangleq[\sigma_1\psi({x}_1'^{T}),\cdots,\sigma_n\psi({x}_n'^{T})]^\top.\]
We have that the random variable $\frac{1}{m}\psi(\tilde{\mathbf{x}}'^{S})G\alpha-\frac{1}{n}\psi(\tilde{\mathbf{x}}'^{T})\mathbf{1}-\frac{1}{m}\psi(\mathbf{x'}^{S})G\alpha+\frac{1}{n}\psi(\mathbf{x'}^{T})\mathbf{1}$ and the random variable $\frac{1}{m}\psi(\tilde{\mathbf{x}}'^{S},\sigma)G\alpha-
\frac{1}{n}\psi(\tilde{\mathbf{x}}'^{T},\sigma)\mathbf{1}-\frac{1}{m}\psi(\mathbf{x'}^{S},\sigma)G\alpha+\frac{1}{n}\psi(\mathbf{x'}^{T},\sigma)\mathbf{1}$ have the same distribution.

Then, we have
\begin{equation}\label{rade3}
\begin{aligned}
&\mathbb{E}_{\mathbf{x}^{S},\mathbf{x}^{T},\tilde{\mathbf{x}}^{S},\tilde{\mathbf{x}}^{T}}\sup_{\alpha\in \Delta}\left\|\left(\frac{1}{m}\psi(\tilde{\mathbf{x}}'^{S})G\alpha-\frac{1}{n}\psi(\tilde{\mathbf{x}}'^{T})\mathbf{1}\right) -\frac{1}{m}\psi(\mathbf{x'}^{S})G\alpha+\frac{1}{n}\psi(\mathbf{x'}^{T})\mathbf{1}\right\|\nonumber\\
&=\mathbb{E}_{\mathbf{x}^{S},\mathbf{x}^{T},\tilde{\mathbf{x}}^{S},\tilde{\mathbf{x}}^{T},\sigma}\sup_{\alpha\in \Delta}\left\|\left(\frac{1}{m}\psi(\tilde{\mathbf{x}}'^{S},\sigma)G\alpha-\frac{1}{n}\psi(\tilde{\mathbf{x}}'^{T},\sigma)\mathbf{1}\right)-\frac{1}{m}\psi(\mathbf{x'}^{S},\sigma)G\alpha+\frac{1}{n}\psi(\mathbf{x'}^{T},\sigma)\mathbf{1}\right\|\nonumber\\
&\leq 2\mathbb{E}_{\mathbf{x}^{S},\mathbf{x}^{T},\sigma}\sup_{\alpha\in \Delta}\left\|\left(\frac{1}{m}\psi(\mathbf{x'}^{S},\sigma)G\alpha-
\frac{1}{n}\psi(\mathbf{x'}^{T},\sigma)\mathbf{1}\right)\right\|\nonumber\\
&\leq 2\mathbb{E}_{\mathbf{x}^{S},\sigma}\sup_{\alpha\in \Delta}\left\|\frac{1}{m}\psi(\mathbf{x'}^{S},\sigma)G\alpha\right\|+2\mathbb{E}_{\mathbf{x}^{T},\sigma}\sup_{\alpha\in \Delta}\left\|
\frac{1}{n}\psi(\mathbf{x'}^{T},\sigma)\mathbf{1}\right\|,
\end{aligned}
\end{equation}
where the inequalities hold because of the triangle inequality.

We then upper bound $\mathbb{E}_{\mathbf{x}^{S},\sigma}\sup_{\alpha\in \Delta}\left\|\frac{1}{m}\psi(\mathbf{x'}^{S},\sigma)G\alpha\right\|$ and $\mathbb{E}_{\mathbf{x}^{T},\sigma}\left\|
\frac{1}{n}\psi(\mathbf{x'}^{T},\sigma)\mathbf{1}\right\|$, respectively. For example, we have
\begin{equation}\label{rade4}
\begin{aligned}
&\mathbb{E}_{\mathbf{x}^{S},\sigma}\sup_{\alpha\in \Delta}\left\|\frac{1}{m}\psi(\mathbf{x'}^{S},\sigma)G\alpha\right\|\nonumber\\
&=\mathbb{E}_{\mathbf{x}^{S},\sigma}\sup_{\alpha\in \Delta}\left\|\frac{1}{m}\left<G^\top[\sigma_1\psi(x_1'^{S}),\cdots,\sigma_m\psi(x_m'^{S})]^\top,\alpha\right>\right\|\nonumber\\
&\leq\mathbb{E}_{\mathbf{x}^{S},\sigma}\sup_{\alpha\in \Delta}\frac{1}{m}\|G^\top[\sigma_1\psi(x_1'^{S}),\cdots,\sigma_m\psi(x_m'^{S})]^\top\|\|\alpha\|\nonumber\\
&\leq\mathbb{E}_{\mathbf{x}^{S},\sigma}\sup_{\alpha\in \Delta}\frac{1}{m}\|G^\top[\sigma_1\psi(x_1'^{S}),\cdots,\sigma_m\psi(x_m'^{S})]^\top\|\|\alpha\|_1\nonumber\\
&\leq \mathbb{E}_{\mathbf{x}^{S},\sigma}\frac{1}{m}\|G^\top[\sigma_1\psi(x_1'^{S}),\cdots,\sigma_m\psi(x_m'^{S})]^\top\|\nonumber\\
&\leq\frac{\wedge_{\hat{Q}}\wedge_{\hat{W}}}{m}\mathbb{E}_{\sigma}\sqrt{c(\sum_{i=1}^{m}\sigma_i)^2}\nonumber\\
&\leq\frac{\wedge_{\hat{Q}}\wedge_{\hat{W}}}{m}\sqrt{c\mathbb{E}_{\sigma}(\sum_{i=1}^{m}\sigma_i)^2}\nonumber\\
&=\frac{\wedge_{\hat{Q}}\wedge_{\hat{W}}\sqrt{c}}{\sqrt{m}},
\end{aligned}
\end{equation}
where $G\in \mathbb{R}^{m\times c}$, $c$ is the number of classes. The first inequality holds because of Cauchy-Schwarz inequality. The second inequality holds because $\|\alpha\|\leq\|\alpha\|_1$. The fourth inequality holds because of the Talagrand Contraction Lemma \cite{ledoux2013probability}. And the last inequality holds because of the Jensen's inequality and that the function sqrt is a concave function. Similarly, we can prove that
\begin{equation}\label{rade5}
\mathbb{E}_{\mathbf{x}^{T},\sigma}\left\|
\frac{1}{n}\psi(\mathbf{x'}^{T},\sigma)\mathbf{1}\right\|\leq\frac{\wedge_{\hat{W}}}{\sqrt{n}}.
\end{equation}

Combining Eq. (\ref{rade1}), Eq. (\ref{rade2}), Eq. (\ref{rade3}), Eq. (\ref{rade4}), and Eq. (\ref{rade5}), we have
\begin{equation}
\begin{aligned}
&\mathbb{E}\sup_{\alpha\in \Delta}\|f(\mathbf{x}^{S},\mathbf{x}^{T},\alpha)\|^2 \\
& \leq4(\wedge_{\hat{Q}}+1)\wedge_{\hat{W}}(\frac{\wedge_{\hat{Q}}\wedge_{\hat{W}}\sqrt{c}}{\sqrt{m}}+\frac{\wedge_{\hat{W}}}{\sqrt{n}})\\
&\leq4(\wedge_{\hat{Q}}+1)^2\wedge_{\hat{W}}^2\sqrt{c}(\frac{1}{\sqrt{m}}+\frac{1}{\sqrt{n}}).
\end{aligned}
\end{equation}
The proof of Lemma \ref{le2} ends. \hfill$\blacksquare$

Now, we are ready to prove Theorem 2.

$\textbf{Proof of Theorem 2.}$
According to Lemma \ref{le1}, we have
\begin{equation}
\begin{aligned}
\mathcal{D}(\hat{W},\hat{\alpha})-\mathcal{D}(\hat{W},\alpha^*) &\leq 2\sup_{\alpha\in \Delta}|\mathcal{D}(\hat{W},{\alpha})-\hat{\mathcal{D}}(\hat{W},\alpha)| \\
&\leq 4(\wedge_{\hat{Q}}+1)\wedge_{\hat{W}}\sup_{\alpha\in \Delta}\|f(\mathbf{x}^{S},\mathbf{x}^{T},\alpha)\|.
\end{aligned}
\end{equation}

Since $\|f(\mathbf{x}^{S},\mathbf{x}^{T},\alpha)\|\geq0$, it holds that 
\begin{equation}
\sup_{\alpha\in\Delta}\|f(\mathbf{x}^{S},\mathbf{x}^{T},\alpha)\|=\sqrt{\sup_{\alpha\in\Delta}\|f(\mathbf{x}^{S},\mathbf{x}^{T},\alpha)\|^2}.
\end{equation}

Then, we will employ McDiarmid's inequality to upper bound the defect $\sup_{\alpha\in\Delta}\|f(\mathbf{x}^{S},\mathbf{x}^{T},\alpha)\|^2$.  We now check its bounded difference property.

Let $\mathbf{x}^{Si}$ be a new sample in the source domain with the $i$-th example in $\mathbf{x}^{S}$ being replaced by an independent example $\tilde{x}_i^{S}$, where $i\in\{1,\cdots,m\}$, and $\mathbf{x}^{Ti}$ be a new sample in the target domain with the $i$-th example in $\mathbf{x}^{T}$ being replaced by an independent example $\tilde{x}_i^{T}$, where $i\in\{1,\cdots,n\}$.

For any $i\in\{1,\cdots,m\}$, we have
\begin{equation}
\begin{aligned}
&\left|\sup_{\alpha\in\Delta}\|f(\mathbf{x}^{Si},\mathbf{x}^{T},\alpha)\|^2-\sup_{\alpha\in\Delta}\|f(\mathbf{x}^{S},\mathbf{x}^{T},\alpha)\|^2\right|\\
&\leq\sup_{\alpha\in\Delta}\left|(f(\mathbf{x}^{Si},\mathbf{x}^{T},\alpha)+f(\mathbf{x}^{S},\mathbf{x}^{T},\alpha))^\top \left(f(\mathbf{x}^{Si},\mathbf{x}^{T},\alpha)-f(\mathbf{x}^{S},\mathbf{x}^{T},\alpha)\right)\right|\\
&\leq\sup_{\alpha\in\Delta}\left|4(\wedge_{\hat{Q}}+1)\psi_{\max}^\top\left(f(\mathbf{x}^{Si},\mathbf{x}^{T},\alpha)-f(\mathbf{x}^{S},\mathbf{x}^{T},\alpha)\right)\right|\\
&=\sup_{\alpha\in\Delta}\left|4(\wedge_{\hat{Q}}+1)\psi_{\max}^\top\left(\frac{1}{m}\psi(\mathbf{x'}^{Si})G\alpha-\frac{1}{m}\psi(\mathbf{x'}^{S})G\alpha\right)\right|\\
&\leq \frac{8\wedge_{\hat{Q}}(\wedge_{\hat{Q}}+1)}{m}|\psi_\text{max}|^\top|\psi_\text{max}|\\
&\leq \frac{8(\wedge_{\hat{Q}}+1)^2\wedge_{\hat{W}}^2}{m}.
\end{aligned}
\end{equation}
Similarly, for any $i\in\{1,\cdots,n\}$, we have
\begin{equation}
\begin{aligned}
&\left|\sup_{\alpha\in\Delta}\|f(\mathbf{x}^{S},\mathbf{x}^{Ti},\alpha)\|^2-\sup_{\alpha\in\Delta}\|f(\mathbf{x}^{S},\mathbf{x}^{T},\alpha)\|^2\right|\\
&\leq\sup_{\alpha\in\Delta}\left|\left(f(\mathbf{x}^{S},\mathbf{x}^{Ti},\alpha)+f(\mathbf{x}^{S},\mathbf{x}^{T},\alpha)\right)^\top \left(f(\mathbf{x}^{S},\mathbf{x}^{Ti},\alpha)-f(\mathbf{x}^{S},\mathbf{x}^{T},\alpha)\right)\right|\\
&\leq\sup_{\alpha\in\Delta}\left|4(\wedge_{\hat{Q}}+1)\psi_{\max}^\top\left(f(\mathbf{x}^{S},\mathbf{x}^{Ti},\alpha)-f(\mathbf{x}^{S},\mathbf{x}^{T},\alpha)\right)\right|\\
&=\sup_{\alpha\in\Delta}\left|4(\wedge_{\hat{Q}}+1)\psi_{\max}^\top\left(\frac{1}{n}\psi(\mathbf{x'}^{Ti})\mathbf{1}-\frac{1}{n}\psi(\mathbf{x'}^{T})\mathbf{1}\right)\right|\\
&\leq \frac{8(\wedge_{\hat{Q}}+1)}{n}|\psi_\text{max}|^\top|\psi_\text{max}|\\
&\leq \frac{8(\wedge_{\hat{Q}}+1)\wedge_{\hat{W}}^2}{n}.
\end{aligned}
\end{equation}

Employing McDiarmid's inequality, we have that
\begin{equation}
\begin{aligned}
&P(\sup_{\alpha\in\Delta}\|f(\mathbf{x}^{S},\mathbf{x}^{T},\alpha)\|^2-\mathbb{E}_{{\bf x}^\mathcal{S},{\bf x}^\mathcal{T}}\sup_{\alpha\in\Delta}\|f(\mathbf{x}^{S},\mathbf{x}^{T},\alpha)\|^2\geq{\epsilon})\\
&\leq\exp\left(\frac{-{\epsilon}^2}{32(\wedge_{\hat{Q}}+1)^4\wedge_{\hat{W}}^4(\frac{1}{m}+\frac{1}{n})}\right).
\end{aligned}
\end{equation}
Let
\[\delta=\exp\left(\frac{-{\epsilon}^2}{32(\wedge_{\hat{Q}}+1)^4\wedge_{\hat{W}}^4(\frac{1}{m}+\frac{1}{n})}\right).\]
For any $\delta>0$, with probability at least $1-\delta$, we have
\begin{equation}
\begin{aligned}
&\sup_{\alpha\in\Delta}\|f(\mathbf{x}^{S},\mathbf{x}^{T},\alpha)\|\\
&\leq\sqrt{\mathbb{E}\sup_{\alpha\in\Delta}\|f(\mathbf{x}^{S},\mathbf{x}^{T},\alpha)\|^2+8(\wedge_{\hat{Q}}+1)^2\wedge_{\hat{W}}^2\sqrt{\frac{1}{2}(\frac{1}{m}+\frac{1}{n})\log\frac{1}{\delta}}}.\\
&\leq (\wedge_{\hat{Q}}+1)\wedge_{\hat{W}} \sqrt{4\sqrt{c}(\frac{1}{\sqrt{m}}+\frac{1}{\sqrt{n}})+\sqrt{32(\frac{1}{m}+\frac{1}{n})\log\frac{1}{\delta}}}\\
\end{aligned}
\end{equation}

Combining the above inequality with those in Lemma \ref{le1} and Lemma \ref{le2}, we have
\begin{equation}
\begin{aligned}
&\mathcal{D}(\hat{W},\hat{\alpha})-\mathcal{D}(\hat{W},\alpha^*)\\
&\leq 2\sup_{\alpha\in \Delta}|\mathcal{D}(\hat{W},{\alpha})-\hat{\mathcal{D}}(\hat{W},\alpha)|\\
&\leq 4(\wedge_{\hat{Q}}+1)\wedge_{\hat{W}}\sup_{\alpha\in \Delta}\|f(\mathbf{x}^{S},\mathbf{x}^{T},\alpha)\|\\
&\leq8(\wedge_{\hat{Q}}+1)^2\wedge_{\hat{W}}^2\sqrt{\frac{\sqrt{c}}{\sqrt{m}}+\frac{\sqrt{c}}{\sqrt{n}}+\sqrt{2(\frac{1}{m}+\frac{1}{n})\log\frac{1}{\delta}}},
\end{aligned}
\end{equation}
which concludes the proof of Theorem 2. \hfill$\blacksquare$
\newpage

\bibliography{egbib}

\begin{thebibliography}{10}

\bibitem{azadi2015auxiliary}
Samaneh Azadi, Jiashi Feng, Stefanie Jegelka, and Trevor Darrell.
\newblock Auxiliary image regularization for deep {CNN}s with noisy labels.
\newblock In {\em ICLR}, 2016.

\bibitem{baktashmotlagh2013unsupervised}
Mahsa Baktashmotlagh, Mehrtash~T Harandi, Brian~C Lovell, and Mathieu Salzmann.
\newblock Unsupervised domain adaptation by domain invariant projection.
\newblock In {\em CVPR}, pages 769--776, 2013.

\bibitem{bartlett2002rademacher}
Peter~L Bartlett and Shahar Mendelson.
\newblock Rademacher and gaussian complexities: {R}isk bounds and structural
  results.
\newblock {\em Journal of Machine Learning Research}, 3(Nov):463--482, 2002.

\bibitem{boucheron2013concentration}
St{\'e}phane Boucheron, G{\'a}bor Lugosi, and Pascal Massart.
\newblock {\em Concentration inequalities: {A} nonasymptotic theory of
  independence.}
\newblock Oxford University Press, 2013.

\bibitem{dubois2017effectiveness}
Sebastien Dubois, Nathanael Romano, Kenneth Jung, Nigam Shah, and David~C Kale.
\newblock The effectiveness of transfer learning in electronic health records
  data.
\newblock In {\em ICLR Workship Track}, 2017.

\bibitem{gong2016domain}
Mingming Gong, Kun Zhang, Tongliang Liu, Dacheng Tao, Clark Glymour, and
  Bernhard Sch{\"o}lkopf.
\newblock Domain adaptation with conditional transferable components.
\newblock In {\em ICML}, pages 2839--2848, 2016.

\bibitem{gopalan2011domain}
Raghuraman Gopalan, Ruonan Li, and Rama Chellappa.
\newblock Domain adaptation for object recognition: {A}n unsupervised approach.
\newblock In {\em ICCV}, pages 999--1006. IEEE, 2011.

\bibitem{hoffman2016fcns}
Judy Hoffman, Dequan Wang, Fisher Yu, and Trevor Darrell.
\newblock {FCNs} in the wild: Pixel-level adversarial and constraint-based
  adaptation.
\newblock {\em arXiv preprint arXiv:1612.02649}, 2016.

\bibitem{huang2007correcting}
Jiayuan Huang, Arthur Gretton, Karsten~M Borgwardt, Bernhard Sch{\"o}lkopf, and
  Alex~J Smola.
\newblock Correcting sample selection bias by unlabeled data.
\newblock In {\em NIPS}, pages 601--608, 2007.

\bibitem{iyer2014maximum}
Arun Iyer, J~Saketha Nath, and Sunita Sarawagi.
\newblock Maximum {M}ean {D}iscrepancy for class ratio estimation:
  {C}onvergence bounds and kernel selection.
\newblock In {\em ICML}, pages 530--538, 2014.

\bibitem{jia2014caffe}
Yangqing Jia, Evan Shelhamer, Jeff Donahue, Sergey Karayev, Jonathan Long, Ross
  Girshick, Sergio Guadarrama, and Trevor Darrell.
\newblock Caffe: Convolutional architecture for fast feature embedding.
\newblock In {\em The 22nd ACM international conference on Multimedia (ACMMM)},
  pages 675--678. ACM, 2014.

\bibitem{khardon2007noise}
Roni Khardon and Gabriel Wachman.
\newblock Noise tolerant variants of the perceptron algorithm.
\newblock {\em Journal of Machine Learning Research}, 8(Feb):227--248, 2007.

\bibitem{krizhevsky2012imagenet}
Alex Krizhevsky, Ilya Sutskever, and Geoffrey~E Hinton.
\newblock Imagenet classification with deep convolutional neural networks.
\newblock In {\em NIPS}, pages 1097--1105, 2012.

\bibitem{lecun1998gradient}
Yann LeCun, L{\'e}on Bottou, Yoshua Bengio, and Patrick Haffner.
\newblock Gradient-based learning applied to document recognition.
\newblock {\em Proceedings of the IEEE}, 86(11):2278--2324, 1998.

\bibitem{ledoux2013probability}
Michel Ledoux and Michel Talagrand.
\newblock {\em Probability in Banach Spaces: {I}soperimetry and processes.}
\newblock Springer Science \& Business Media, 2013.

\bibitem{lee2017cleannet}
Kuang-Huei Lee, Xiaodong He, Lei Zhang, and Linjun Yang.
\newblock Cleannet: Transfer learning for scalable image classifier training
  with label noise.
\newblock In {\em CVPR}, 2018.

\bibitem{liu2016classification}
Tongliang Liu and Dacheng Tao.
\newblock Classification with noisy labels by importance reweighting.
\newblock {\em IEEE Transactions on Pattern Analysis and Machine Intelligence},
  38(3):447--461, 2016.

\bibitem{long2015learning}
Mingsheng Long, Yue Cao, Jianmin Wang, and Michael~I Jordan.
\newblock Learning transferable features with deep adaptation networks.
\newblock In {\em ICML}, pages 97--105, 2015.

\bibitem{long2016deep}
Mingsheng Long, Jianmin Wang, and Michael~I Jordan.
\newblock Deep transfer learning with joint adaptation networks.
\newblock In {\em ICML}, 2017.

\bibitem{long2008random}
Philip~M Long and Rocco~A Servedio.
\newblock Random classification noise defeats all convex potential boosters.
\newblock In {\em ICML}, pages 608--615, 2008.

\bibitem{mohri2012foundations}
Mehryar Mohri, Afshin Rostamizadeh, and Ameet Talwalkar.
\newblock {\em Foundations of machine learning.}
\newblock MIT Press, 2012.

\bibitem{natarajan2013learning}
Nagarajan Natarajan, Inderjit~S Dhillon, Pradeep~K Ravikumar, and Ambuj Tewari.
\newblock Learning with noisy labels.
\newblock In {\em NIPS}, pages 1196--1204, 2013.

\bibitem{pan2011domain}
Sinno~Jialin Pan, Ivor~W Tsang, James~T Kwok, and Qiang Yang.
\newblock Domain adaptation via transfer component analysis.
\newblock {\em IEEE Transactions on Neural Networks}, 22(2):199--210, 2011.

\bibitem{patrini2017making}
Giorgio Patrini, Alessandro Rozza, Aditya~Krishna Menon, Richard Nock, and
  Lizhen Qu.
\newblock Making deep neural networks robust to label noise: {A} loss
  correction approach.
\newblock In {\em CVPR}, 2017.

\bibitem{pearl2011transportability}
Judea Pearl and Elias Bareinboim.
\newblock Transportability of causal and statistical relations: {A} formal
  approach.
\newblock In {\em AISTATS}, pages 247--254, 2011.

\bibitem{peng2017visda}
Xingchao Peng, Ben Usman, Neela Kaushik, Judy Hoffman, Dequan Wang, and Kate
  Saenko.
\newblock Visda: The visual domain adaptation challenge.
\newblock {\em arXiv preprint arXiv:1710.06924}, 2017.

\bibitem{purushotham2016variational}
Sanjay Purushotham, Wilka Carvalho, Tanachat Nilanon, and Yan Liu.
\newblock Variational recurrent adversarial deep domain adaptation.
\newblock In {\em ICLR}, 2017.

\bibitem{ramaswamy2016mixture}
Harish Ramaswamy, Clayton Scott, and Ambuj Tewari.
\newblock Mixture {P}roportion {E}stimation via kernel embeddings of
  distributions.
\newblock In {\em ICML}, pages 2052--2060, 2016.

\bibitem{reid2010composite}
Mark~D Reid and Robert~C Williamson.
\newblock Composite binary losses.
\newblock {\em Journal of Machine Learning Research}, 11(Sep):2387--2422, 2010.

\bibitem{saez2016influence}
Jos{\'e}~A S{\'a}ez, Bartosz Krawczyk, and Micha{\l} Wo{\'z}niak.
\newblock On the influence of class noise in medical data classification:
  Treatment using noise filtering methods.
\newblock {\em Applied Artificial Intelligence}, 30(6):590--609, 2016.

\bibitem{Scholkopf12}
B.~Sch{\"o}lkopf, D.~Janzing, J.~Peters, E.~Sgouritsa, K.~Zhang, and J.~Mooij.
\newblock On causal and anticausal learning.
\newblock In {\em ICML}, 2012.

\bibitem{scholkopf2012causal}
Bernhard Sch{\"o}lkopf, Dominik Janzing, Jonas Peters, Eleni Sgouritsa, Kun
  Zhang, and Joris Mooij.
\newblock On causal and anticausal learning.
\newblock {\em arXiv preprint arXiv:1206.6471}, 2012.

\bibitem{scott2015rate}
Clayton Scott.
\newblock A rate of convergence for {M}ixture {P}roportion {E}stimation, with
  application to learning from noisy labels.
\newblock In {\em AISTATS}, pages 838--846, 2015.

\bibitem{sukhbaatar2014training}
Sainbayar Sukhbaatar, Joan Bruna, Manohar Paluri, Lubomir Bourdev, and Rob
  Fergus.
\newblock Training convolutional networks with noisy labels.
\newblock {\em arXiv preprint arXiv:1406.2080}, 2014.

\bibitem{torralba2011unbiased}
Antonio Torralba and Alexei~A Efros.
\newblock Unbiased look at dataset bias.
\newblock In {\em CVPR}, pages 1521--1528. IEEE, 2011.

\bibitem{van2015learning}
Brendan Van~Rooyen, Aditya Menon, and Robert~C Williamson.
\newblock Learning with symmetric label noise: {T}he importance of being
  unhinged.
\newblock In {\em NIPS}, pages 10--18, 2015.

\bibitem{wang2014active}
Xuezhi Wang, Tzu-Kuo Huang, and Jeff~G. Schneider.
\newblock Active transfer learning under model shift.
\newblock In {\em ICML}, pages 1305--1313, 2014.

\bibitem{yu2018efficient}
Xiyu Yu, Tongliang Liu, Mingming Gong, Kayhan Batmanghelich, and Dacheng Tao.
\newblock An efficient and provable approach for mixture proportion estimation
  using linear independence assumption.
\newblock In {\em CVPR}, 2018.

\bibitem{zhang2013covariate}
Kai Zhang, Vincent Zheng, Qiaojun Wang, James Kwok, Qiang Yang, and Ivan
  Marsic.
\newblock Covariate shift in hilbert space: A solution via sorrogate kernels.
\newblock In {\em ICML}, pages 388--395, 2013.

\bibitem{zhang2013domain}
Kun Zhang, Bernhard Sch{\"o}lkopf, Krikamol Muandet, and Zhikun Wang.
\newblock Domain adaptation under target and conditional shift.
\newblock In {\em ICML}, pages 819--827, 2013.

\end{thebibliography}
\end{document}